\documentclass[
journal=jacsat, 
manuscript=article]{achemso}

\usepackage[version=3]{mhchem} 
\usepackage{multirow, bm, threeparttable}

\newcommand{\tabincell}[2]{\begin{tabular}{@{}#1@{}}#2\end{tabular}}

\author{Zhenqin Wu}
\affiliation{Department of Chemistry, Stanford University}
\altaffiliation{Joint First Authorship}
\author{Bharath Ramsundar}
\affiliation{Department of Computer Science, Stanford University}
\altaffiliation{Joint First Authorship}
\author{Evan N. Feinberg}
\affiliation{Program in Biophysics, Stanford School of Medicine}
\altaffiliation{Joint Second Authorship}
\author{Joseph Gomes}
\affiliation{Department of Chemistry, Stanford University}
\altaffiliation{Joint Second Authorship}
\author{Caleb Geniesse}
\affiliation{Program in Biophysics, Stanford School of Medicine}
\author{Aneesh S. Pappu}
\affiliation{Department of Computer Science, Stanford University}
\author{Karl Leswing}
\affiliation{Schrodinger Inc.}
\author{Vijay Pande}
\affiliation{Department of Chemistry, Stanford University}
\email{pande@stanford.edu}

\title[\texttt{achemso} demonstration]
{MoleculeNet: A Benchmark for Molecular Machine Learning}
\begin{document}

\begin{abstract}
Molecular machine learning has been maturing rapidly over the last few years. Improved methods and the presence of larger datasets have enabled machine learning algorithms to make increasingly accurate predictions about molecular properties. However, algorithmic progress has been limited due to the lack of a standard benchmark to compare the efficacy of proposed methods; most new algorithms are benchmarked on different datasets making it challenging to gauge the quality of proposed methods. This work introduces MoleculeNet, a large scale benchmark for molecular machine learning. MoleculeNet curates multiple public datasets, establishes metrics for evaluation, and offers high quality open-source implementations of multiple previously proposed molecular featurization and learning algorithms (released as part of the DeepChem open source library). MoleculeNet benchmarks demonstrate that learnable representations are powerful tools for molecular machine learning and broadly offer the best performance. However, this result comes with caveats. Learnable representations still struggle to deal with complex tasks under data scarcity and highly imbalanced classification. For quantum mechanical and biophysical datasets, the use of physics-aware featurizations can be more important than choice of particular learning algorithm.
\end{abstract}

\section{Introduction} \label{Introduction}

Overlap between chemistry and 
statistical learning has had a long history. The field of cheminformatics has been utilizing machine learning methods in chemical modeling(e.g. quantitative structure activity relationships, QSAR) for decades.\cite{gasteiger1993neural, zupan1999neural, varnek2012machine, mitchell2014machine,devillers1996neural, schneider1998artificial} In the recent 10 years, with the advent of sophisticated deep learning methods \cite{DeepLearning, schmidhuber2015deep}, machine learning has gathered increasing amounts of attention from the scientific community. Data-driven analysis has become a routine step in many chemical and biological applications, including virtual screening\cite{ma2015deep, ramsundar2015massively, unterthiner2014deep, AtomNet}, chemical property prediction\cite{ESOL_dataset, lusci2013deep, SAMPL4, FreeSolv}, and quantum chemistry calculations\cite{GDB7_dataset_prl, GDB7_dataset_arxiv, schutt2016quantum, mcgibbon2017improving}. 

In many such applications, machine learning has shown strong potential to compete with or even outperform conventional \textit{ab-initio} computations\cite{FreeSolv, GDB7_dataset_arxiv}. It follows that introduction of novel machine learning methods has the potential to reshape research on properties of molecules. However, this potential has been limited by the lack of a standard evaluation platform for proposed machine learning algorithms. Algorithmic papers often benchmark proposed methods on disjoint dataset collections, making it a challenge to gauge whether a proposed technique does in fact improve performance.

Data for molecule-based machine learning tasks are highly heterogeneous and expensive to gather. Obtaining precise and accurate results for chemical properties typically requires specialized instruments as well as expert supervision (contrast with computer speech and vision, where lightly trained workers can annotate data suitable for machine learning systems). As a result, molecular datasets are usually much smaller than those available for other machine learning tasks. Furthermore, the breadth of chemical research means our interests with respect to a molecule may range from quantum characteristics to measured impacts on the human body. Molecular machine learning methods have to be capable of learning to predict this very broad range of properties. Complicating this challenge, input molecules can have arbitrary size and components, highly variable connectivity and many three dimensional conformers (three dimensional molecular shapes). To transform molecules into a form suitable for conventional machine learning algorithms (that usually accept fixed length input), we have to extract useful and related information from a molecule into a fixed dimensional representation (a process called featurization).\cite{ECFP, graphconv_feat, kearnes2016graphconv}.  

To put it simply, building machine learning models on molecules requires overcoming several key issues: limited amounts of data, wide ranges of outputs to predict, large heterogeneity in input molecular structures and appropriate learning algorithms. Therefore, this work aims to facilitate the development of molecular machine learning methods by curating a number of dataset collections, creating a suite of software that implements many known featurizations of molecules, and providing high quality implementations of many previously proposed algorithms. Following the footsteps of WordNet\cite{wordnet} and ImageNet\cite{imagenet_cvpr09}, we call our suite MoleculeNet, a benchmark collection for molecular machine learning.

In machine learning, a benchmark serves as more than a simple collection of data and methods. The introduction of the ImageNet benchmark in 2009 has triggered a series of breakthroughs in computer vision, and in particular has facilitated the rapid development of deep convolutional networks. The ILSVRC, an annual contest held by the ImageNet team\cite{ILSVRC15}, draws considerable attention from the community, and greatly stimulates collaborations and competitions across the field. The contest has given rise to a series of prominent machine learning models such as AlexNet\cite{alexnet}, GoogLeNet\cite{googlenet}, ResNet\cite{resnet} which have had broad impact on the academic and industrial computer science communities. We hope that MoleculeNet will trigger similar breakthroughs by serving as a platform for the wider community to develop and improve models for learning molecular properties.

In particular, MoleculeNet contains data on the properties of over 700,000 compounds. All datasets have been curated and integrated into the open source DeepChem package.\cite{deepchem} Users of DeepChem can easily load all MoleculeNet benchmark data through provided library calls. MoleculeNet also contributes high quality implementations of well known (bio)chemical featurization methods. To facilitate comparison and development of new methods, we also provide high quality implementations of several previously proposed machine learning methods. Our implementations are integrated with DeepChem, and depend on Scikit-Learn \cite{pedregosa2011scikit} and Tensorflow \cite{abadi2016tensorflow} underneath the hood. Finally, evaluation of machine learning algorithms requires defined methods to split datasets into training/validation/test collections. Random splitting, common in machine learning, is often not correct for chemical data \cite{sheridan2013time}. MoleculeNet contributes a library of splitting mechanisms to DeepChem and evaluates all algorithms with multiple choices of data split. MoleculeNet provide a series of benchmark results of implemented machine learning algorithms using various featurizations and splits upon our dataset collections. These results are provided within this paper, and will be maintained online in an ongoing fashion as part of DeepChem. 

The related work section will review prior work in the chemistry community on gathering curated datasets and discuss how MoleculeNet differs from these previous efforts. The methods section reviews the dataset collections, metrics, featurization methods, and machine learning models included as part of MoleculeNet. The results section will analyze the benchmarking results to draw conclusions about the algorithms and datasets considered.

\section{Related Work}
MoleculeNet draws upon a broader movement within the chemical community to gather large sources of curated data. PubChem \cite{bolton2008pubchem} and PubChem BioAssasy \cite{pcba_dataset} gather together thousands of bioassay results, along with millions of unique molecules tested within these assays. The ChEMBL database offers a similar service, with millions of bioactivity outcomes across thousands of protein targets. Both PubChem and ChEMBL are human researcher oriented, with web portals that facilitate browsing of the available targets and compounds. ChemSpider is a repository of nearly 60 million chemical structures, with web based search capabilities for users. The Crystallography Open Database \cite{gravzulis2009crystallography} and Cambridge Structural Database \cite{groom2016cambridge} offer large repositories of organic and inorganic compounds. The protein data bank \cite{berman2003announcing} offers a repository of experimentally resolved three dimensional protein structures. This listing is by no means comprehensive; the methods section will discuss a number of smaller data sources in greater detail.

These past efforts have been critical in enabling the growth of computational chemistry. However, these previous databases are not machine-learning focused. In particular, these collections don't define metrics which measure the effectiveness of algorithmic methods in understanding the data contained. Furthermore, there is no prescribed separation of the data into training/validation/test sets (critical for machine learning development). Without specified metrics or splits, the choice is left to individual researchers, and there are indeed many chemical machine learning papers which use subsets of these data stores for machine learning evaluation. Unfortunately, the choice of metric and subset varies widely between groups, so two methods papers using PubChem data may be entirely incomparable. MoleculeNet aims to bridge this gap by providing benchmark results for a reasonable range of metrics, splits, and subsets of these (and other) data collections.

It's important to note that there have been some efforts to create benchmarking datasets for machine learning in chemistry. The Quantum Machine group \cite{QuantumMachine} and previous work on multitask learning \cite{ramsundar2015massively} both introduce benchmarking collections which have been used in multiple papers. MoleculeNet incorporates data from both these efforts and significantly expands upon them.

\section{Methods}

\begin{figure}
  \includegraphics[width=\textwidth]{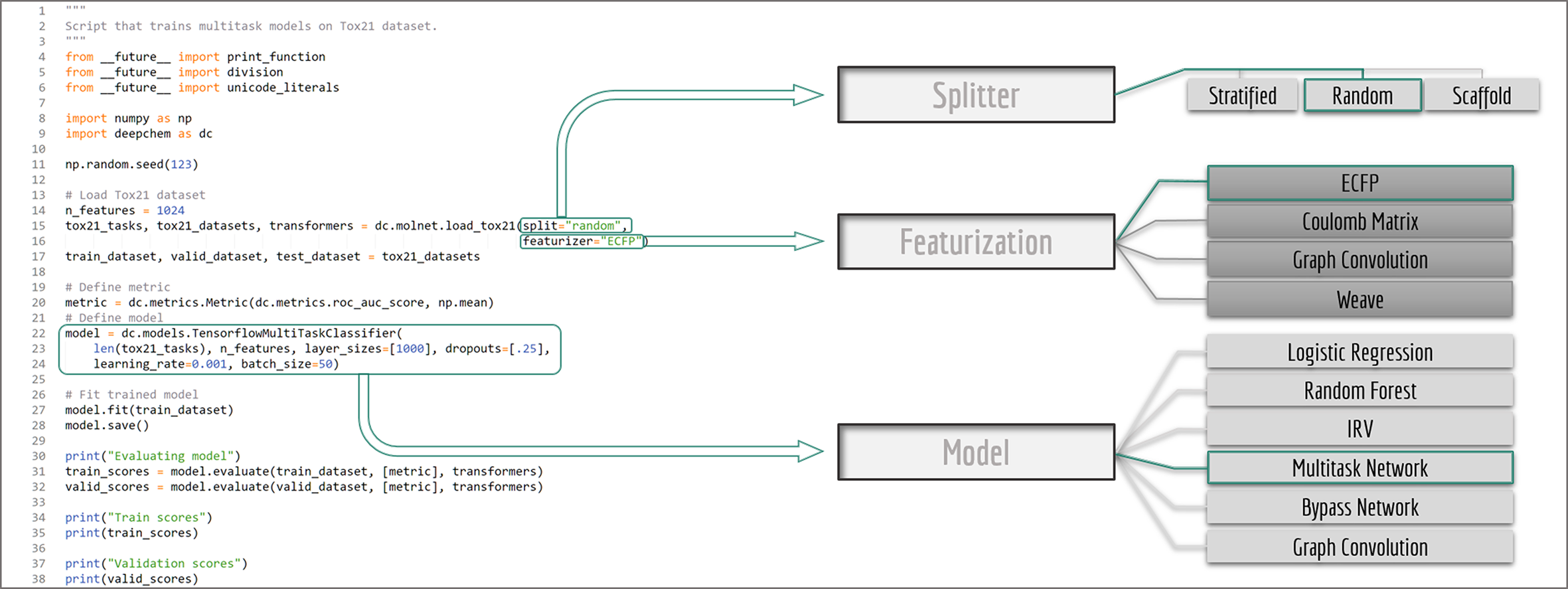}
  \caption{Example code for benchmark evaluation with DeepChem, multiple methods are provided for data splitting, featurization and learning.}
  \label{fig:benchmark_example}
\end{figure}

MoleculeNet is based on the open source package DeepChem\cite{deepchem}. Figure~\ref{fig:benchmark_example} shows an annotated DeepChem benchmark script. Note how different choices for data splitting, featurization, and model are available. DeepChem also directly provides molnet sub-module to support benchmarking. The single line below runs benchmarking on the specified dataset, model and featurizer. User defined models capable of handling DeepChem datasets are also supported.

{\fontfamily{pcr}\selectfont
deepchem.molnet.run\_benchmark(datasets, model, split, featurizer)
}

In this section, we will further elaborate the benchmarking system, introducing available datasets as well as implemented splitting, metrics, featurization, and learning methods.

\subsection{Datasets}
\begin{figure}[H]
  \centering
  \includegraphics[width=\textwidth]{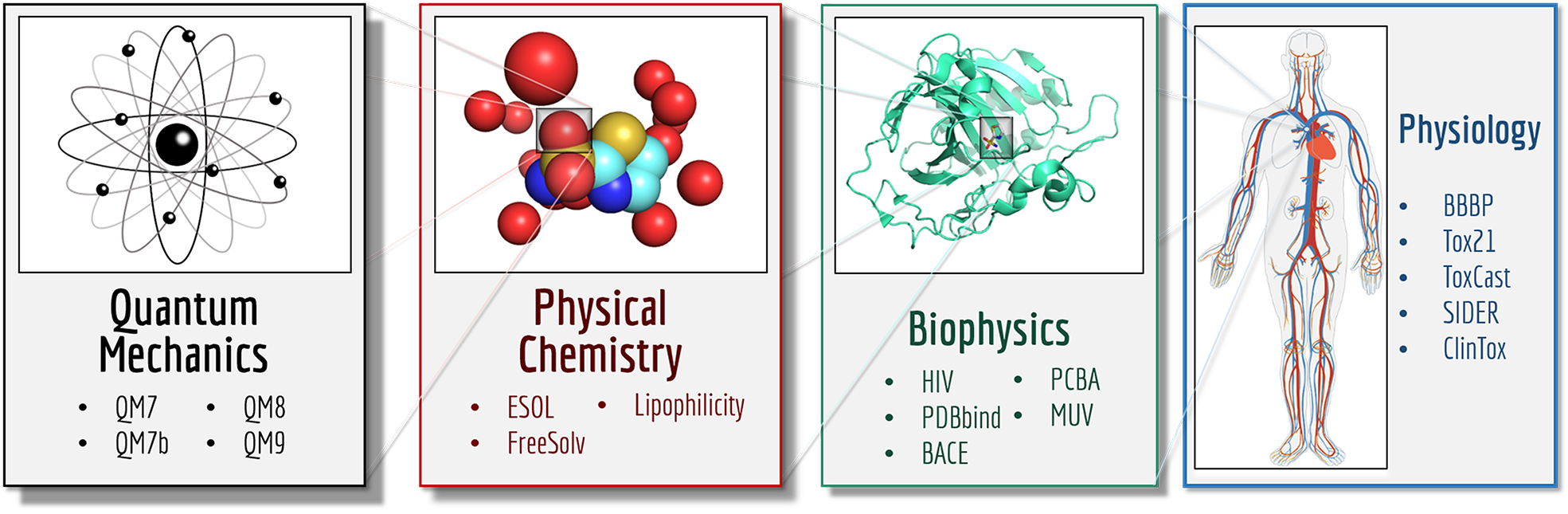}
  \caption{Tasks in different datasets focus on different levels of properties of molecules.}
  \label{fig:dataset_composition}
\end{figure}

MoleculeNet is built upon multiple public databases. The full collection currently includes over 700,000 compounds tested on a range of different properties. These properties can be subdivided into four categories: quantum mechanics, physical chemistry, biophysics and physiology. As illustrated in Figure~\ref{fig:dataset_composition}, separate datasets in the MoleculeNet collection cover various levels of molecular properties, ranging from molecular-level properties to macroscopic influences on human body. For each dataset, we propose a metric and a splitting pattern(introduced in the following texts) that best fit the properties of the dataset. Performances on the recommended metric and split are reported in the results section.

In most datasets, SMILES strings\cite{SMILES} are used to represent input molecules, 3D coordinates are also included in part of the collection as molecular features, which enabled different methods to be applied. Properties, or output labels, are either 0/1 for classification tasks, or floating point numbers for regression tasks. At the time of writing, MoleculeNet contains 17 datasets prepared and benchmarked, but we anticipate adding further datasets in an on-going fashion. We also highly welcome contributions from other public data collections. For more detailed dataset structure requirements and instructions on curating datasets, please refer to the tutorial on DeepChem webpage.

Table~\ref{tab:n_samples} lists details of datasets in the collection, including tasks, compounds and their features, recommended splits and metrics. Contents of each dataset will be elaborated in this subsection.

\begin{table}[h]
    \caption{Dataset Details: number of compounds and tasks, recommended splits and metrics}
    \label{tab:n_samples}
    \centering
    \tiny
    \begin{tabular}{ |c|c|c|c|c|c|c|c| }
    \hline
    \textbf{Category} & \textbf{Dataset} & \textbf{Data Type} & \textbf{\# Tasks} & \textbf{Task Type} & \textbf{\# Compounds} & \textbf{Rec - Split} & \textbf{Rec - Metric}\\ 
    \hline
    \multirow{4}{*}{Quantum Mechanics} & QM7 & SMILES, 3D coordinates & 1 & Regression & 7160 & Stratified & MAE\\\cline{2-8}
    & QM7b & 3D coordinates & 14 & Regression & 7210 & Random & MAE\\\cline{2-8}
    & QM8 & SMILES, 3D coordinates & 12 & Regression & 21786 & Random & MAE \\\cline{2-8}
    & QM9 & SMILES, 3D coordinates & 12 & Regression & 133885 & Random & MAE \\\cline{2-8}
    \hline
    \multirow{3}{*}{Physical Chemistry} & ESOL & SMILES & 1 & Regression & 1128 & Random & RMSE \\\cline{2-8}
    & FreeSolv & SMILES & 1 & Regression & 642 & Random & RMSE \\\cline{2-8}
    & Lipophilicity & SMILES & 1 & Regression & 4200 & Random & RMSE \\
    \hline
    \multirow{5}{*}{Biophysics} & PCBA & SMILES & 128 & Classification & 437929 & Random & PRC-AUC \\\cline{2-8}
    & MUV & SMILES & 17 & Classification & 93087 & Random & PRC-AUC \\\cline{2-8}
    & HIV & SMILES & 1 & Classification & 41127 & Scaffold & ROC-AUC \\\cline{2-8}
    & PDBbind & SMILES, 3D coordinates & 1 & Regression & 11908 & Time & RMSE \\\cline{2-8}
    & BACE & SMILES & 1 & Classification & 1513 & Scaffold & ROC-AUC \\
    \hline
    \multirow{5}{*}{Physiology} & BBBP & SMILES & 1 & Classification & 2039 & Scaffold & ROC-AUC \\\cline{2-8}
    & Tox21 & SMILES & 12 & Classification & 7831 & Random & ROC-AUC \\\cline{2-8}
    & ToxCast & SMILES & 617 & Classification & 8575 & Random & ROC-AUC \\\cline{2-8}
    & SIDER & SMILES & 27 & Classification & 1427 & Random & ROC-AUC \\\cline{2-8}
    & ClinTox & SMILES & 2 & Classification & 1478 & Random & ROC-AUC \\
    \hline
    \end{tabular}
\end{table}
\subsubsection{QM7/QM7b}

The QM7/QM7b datasets are subsets of the GDB-13 database\cite{GDB13}, a database of nearly 1 billion stable and synthetically accessible organic molecules, containing up to seven ``heavy'' atoms (C, N, O, S). The 3D Cartesian coordinates of the most stable conformation and electronic properties (atomization energy, HOMO/LUMO eigenvalues, etc.) of each molecule were determined using \textit{ab-initio} density functional theory (PBE0/tier2 basis set).\cite{GDB7_dataset_prl, GDB7_dataset_arxiv} Learning methods benchmarked on QM7/QM7b are responsible for predicting these electronic properties given stable conformational coordinates. For the purpose of more stable performances as well as better comparison, we recommend stratified splitting(introduced in the next subsection) for QM7.

\subsubsection{QM8}

The QM8 dataset comes from a recent study on modeling quantum mechanical calculations of electronic spectra and excited state energy of small molecules.\cite{QM8} Multiple methods, including time-dependent density functional theories (TDDFT) and second-order approximate coupled-cluster (CC2), are applied to a collection of molecules that include up to eight heavy atoms (also a subset of the GDB-17 database\cite{GDB-17}). In total, four excited state properties are calculated by three different methods on 22 thousand samples.

\subsubsection{QM9}

QM9 is a comprehensive dataset that provides geometric, energetic, electronic and thermodynamic properties for a subset of GDB-17 database\cite{GDB-17}, comprising 134 thousand stable organic molecules with up to nine heavy atoms\cite{QM9}. All molecules are modeled using density functional theory (B3LYP/6-31G(2df,p) based DFT). In our benchmark, geometric properties (atomic coordinates) are integrated into features, which are then applied to predict other properties.

The datasets introduced above (QM7, QM7b, QM8, QM9) were curated as part of the Quantum-Machine effort \cite{QuantumMachine}, which has processed a number of datasets to measure the efficacy of machine-learning methods for quantum chemistry.

\subsubsection{ESOL}

ESOL is a small dataset consisting of water solubility data for 1128 compounds.\cite{ESOL_dataset} The dataset has been used to train models that estimate solubility directly from chemical structures (as encoded in SMILES strings).\cite{graphconv_feat} Note that these structures don't include 3D coordinates, since solubility is a property of a molecule and not of its particular conformers.

\subsubsection{FreeSolv}

The Free Solvation Database (FreeSolv) provides experimental and calculated hydration free energy of small molecules in water\cite{FreeSolv}. A subset of the compounds in the dataset are also used in the SAMPL blind prediction challenge\cite{SAMPL4}. The calculated values are derived from alchemical free energy calculations using molecular dynamics simulations. We include the experimental values in the benchmark collection, and use calculated values for comparison.

\subsubsection{Lipophilicity}

Lipophilicity is an important feature of drug molecules that affects both membrane permeability and solubility. This dataset, curated from ChEMBL database,\cite{Hersey2015lipo} provides experimental results of octanol/water distribution coefficient (logD at pH 7.4) of 4200 compounds.

\subsubsection{PCBA}

PubChem BioAssay (PCBA) is a database consisting of biological activities of small molecules generated by high-throughput screening.\cite{pcba_dataset} We use a subset of PCBA, containing 128 bioassays measured over 400 thousand compounds, used by previous work to benchmark machine learning methods.\cite{ramsundar2015massively}

\subsubsection{MUV}

The Maximum Unbiased Validation (MUV) group is another benchmark dataset selected from PubChem BioAssay by applying a refined nearest neighbor analysis.\cite{muv_dataset} The MUV dataset contains 17 challenging tasks for around 90 thousand compounds and is specifically designed for validation of virtual screening techniques.

\subsubsection{HIV}

The HIV dataset was introduced by the Drug Therapeutics Program (DTP) AIDS Antiviral Screen, which tested the ability to inhibit HIV replication for over 40,000 compounds.\cite{HIV} Screening results were evaluated and placed into three categories: confirmed inactive (CI), confirmed active (CA) and confirmed moderately active (CM). We further combine the latter two labels, making it a classification task between inactive (CI) and active (CA and CM). As we are more interested in discover new categories of HIV inhibitors, scaffold splitting(introduced in the next subsection) is recommended for this dataset.

\subsubsection{PDBbind}

PDBbind is a comprehensive database of experimentally measured binding affinities for bio-molecular complexes.\cite{PDBbind1, PDBbind2} Unlike other ligand-based biological activity datasets, in which only the structures of ligands are provided, PDBbind provides detailed 3D Cartesian coordinates of both ligands and their target proteins derived from experimental (e.g., X-Ray crystallography) measurements. The availability of coordinates of the protein-ligand complexes permits structure-based featurization that is aware of the protein-ligand binding geometry. We use the ``refined'' and ``core'' subsets of the database\cite{PDBbind3}, more carefully processed for data artifacts, as additional benchmarking targets. Samples in PDBbind dataset are collected over a relatively long period of time(since 1982), hence a time splitting pattern(introduced in the next subsection) is recommended to mimic actual development in the field.

\subsubsection{BACE}

The BACE dataset provides quantitative ($IC_{50}$) and qualitative (binary label) binding results for a set of inhibitors of human $\beta$-secretase 1 (BACE-1)\cite{BACE-1}. All data are experimental values reported in scientific literature over the past decade, some with detailed crystal structures available. We merged a collection of 1522 compounds with their 2D structures and binary labels in MoleculeNet, built as a classification task. Similarly, regarding a single protein target, scaffold splitting will be more practically useful.

\subsubsection{BBBP}

The Blood-brain barrier penetration (BBBP) dataset comes from a recent study\cite{BBBP} on the modeling and prediction of the barrier permeability. As a membrane separating circulating blood and brain extracellular fluid, the blood-brain barrier blocks most drugs, hormones and neurotransmitters. Thus penetration of the barrier forms a long-standing issue in development of drugs targeting central nervous system. This dataset includes binary labels for over 2000 compounds on their permeability properties. Scaffold splitting is also recommended for this well-defined target.

\subsubsection{Tox21}

The ``Toxicology in the 21st Century'' (Tox21) initiative created a public database measuring toxicity of compounds, which has been used in the 2014 Tox21 Data Challenge \cite{Tox21}. This dataset contains qualitative toxicity measurements for 8014 compounds on 12 different targets, including nuclear receptors and stress response pathways.

\subsubsection{ToxCast}

ToxCast is another data collection (from the same initiative as Tox21) providing toxicology data for a large library of compounds based on \textit{in vitro} high-throughput screening.\cite{toxcast_dataset} The processed collection in MoleculeNet includes qualitative results of over 600 experiments on 8615 compounds.

\subsubsection{SIDER}

The Side Effect Resource (SIDER) is a database of marketed drugs and adverse drug reactions (ADR) \cite{sider_dataset}. The version of the SIDER dataset in DeepChem \cite{altae2016low} has grouped drug side-effects into 27 system organ classes following MedDRA classifications \cite{meddra} measured for 1427 approved drugs (following previous usage \cite{altae2016low}).

\subsubsection{ClinTox}

The ClinTox dataset, introduced as part of this work, compares drugs approved by the FDA and drugs that have failed clinical trials for toxicity reasons. \cite{PrOCTOR2016, InSilicoMed2016} The dataset includes two classification tasks for 1491 drug compounds with known chemical structures: (1) clinical trial toxicity (or absence of toxicity) and (2) FDA approval status. List of FDA-approved drugs are compiled from the SWEETLEAD database,\cite{SWEETLEAD_database} and list of drugs that failed clinical trials for toxicity reasons are compiled from the Aggregate Analysis of ClinicalTrials.gov (AACT) database.\cite{AACT}

\subsection{Dataset splitting}

\begin{figure}[htbp]
  \includegraphics[width=.5\textwidth]{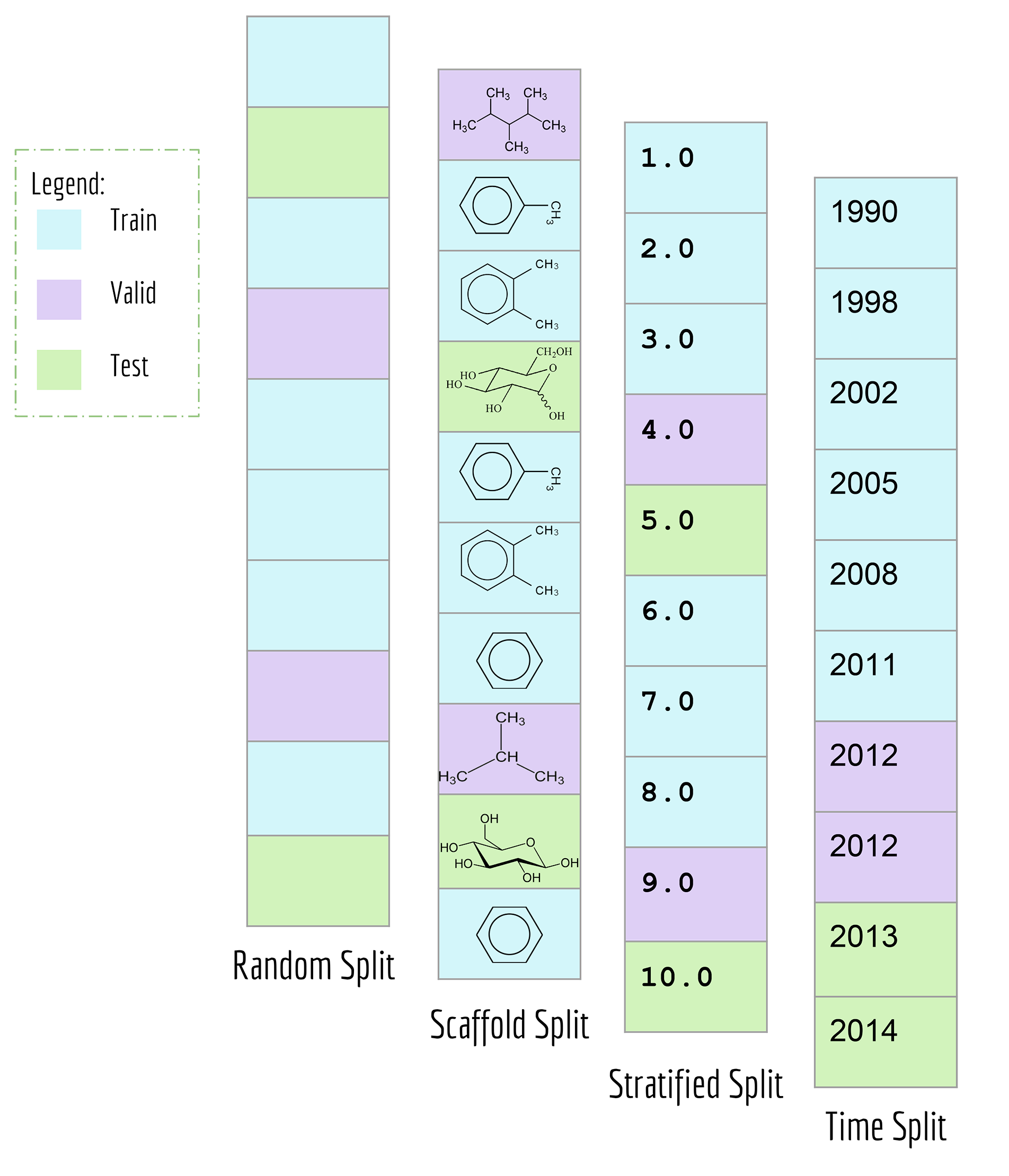}
  \caption{Representation of Data Splits in MoleculeNet.}
  \label{fig:data_splits}
\end{figure}

Typical machine learning methods require datasets to be split into training/validation/test subsets (or alternatively into $K$-folds) for benchmarking. All MoleculeNet datasets are split into training, validation and test, following a 80/10/10 ratio. Training sets were used to train models, while validation sets were used for tuning hyperparameters, and test sets were used for evaluation of models.

As mentioned previously, random splitting of molecular data isn't always best for evaluating machine learning methods. Consequently, MoleculeNet implements multiple different splittings for each dataset. Random splitting randomly splits samples into the training/validation/test subsets. Scaffold splitting splits the samples based on their two-dimensional structural frameworks,\cite{scaffold} as implemented in RDKit.\cite{RDKit} Since scaffold splitting attempts to separate structurally different molecules into different subsets, it offers a greater challenge for learning algorithms than the random split.

In addition, a stratified random sampling method is implemented on the QM7 dataset to reproduce the results from the original work.\cite{GDB7_dataset_arxiv} This method sorts datapoints in order of increasing label value (note this is only defined for real-valued output). This sorted list is then split into training/validation/test by ensuring that each set contains the full range of provided labels. Time splitting is also adopted for dataset that includes time information(PDBbind). Under this splitting method, model will be trained on older data and tested on newer data, mimicking real world development condition.

MoleculeNet contributes the code for these splitting methods into DeepChem. Users of the library can use these splits on new datasets with short library calls.

\subsection{Metrics}

MoleculeNet contains both regression datasets (QM7, QM7b, QM8, QM9, ESOL, FreeSolv, Lipophilicity and PDBbind) and classification datasets (PCBA, MUV, HIV, BACE, BBBP, Tox21, ToxCast and SIDER). Consequently, different performance metrics need to be measured for each. Following suggestions from the community\cite{metric_suggestion}, regression datasets are evaluated by mean absolute error (MAE) and root-mean-square error (RMSE), classification datasets are evaluated by area under curve (AUC) of the receiver operating characteristic (ROC) curve\cite{AUC-ROC} and the precision recall curve (PRC)\cite{PRC}. For datasets containing more than one task, we report the mean metric values over all tasks.

\begin{figure}[h]
  \centering
  \small
  \includegraphics[width=0.5\textwidth]{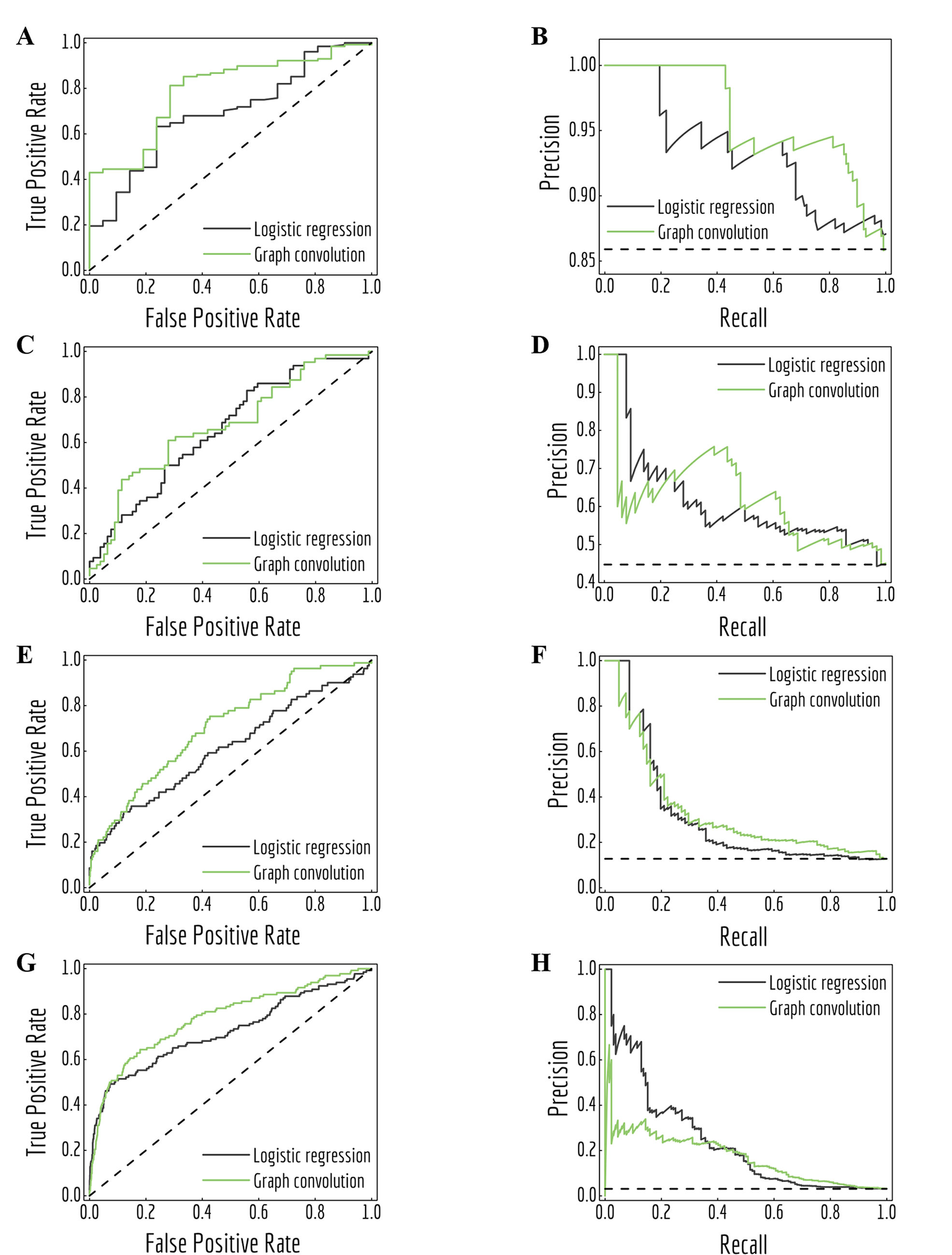}
  \caption{Receiver operating characteristic (ROC) curves and precision recall curves (PRC) for predictions of logistic regression and graph convolutional models under different class imbalance condition.(Details listed in Table~\ref{tab:ROCvsPRC_AUC}): \textbf{A, B}: task "FDA\_APPROVED" from ClinTox, test subset; \textbf{C, D}: task "Hepatobiliary disorders" from SIDER, test subset; \textbf{E, F}: task "NR-ER" from Tox21, validation subset; \textbf{G, H}: task "HIV\_active" from HIV, test subset. Black dashed lines are performances of random classifiers.}
  \label{fig:ROCvsPRC}
\end{figure}

\begin{table}[h]
    \caption{Task details and area under curve(AUC) values of sample curves}
    \label{tab:ROCvsPRC_AUC}
    \centering
    \begin{tabular}{ |c|c|c|c|c| }
    \hline
    \textbf{Task} & \textbf{P/N}* & \textbf{Model} & \textbf{ROC} & \textbf{PRC}\\ 
    \hline
    \multirow{2}{*}{\tabincell{l}{``FDA\_APPROVED''\\ClinTox, test subset}} &  \multirow{2}{*}{128/21} &{\scriptsize Logistic Regression} & 0.691 & 0.932 \\\cline{3-5}
    & & {\scriptsize Graph Convolution} & 0.791 & 0.959 \\
    \hline
    \multirow{2}{*}{\tabincell{l}{``Hepatobiliary disorders''\\SIDER, test subset}} &  \multirow{2}{*}{64/79} & {\scriptsize Logistic Regression} & 0.659 & 0.612 \\\cline{3-5}
    & & {\scriptsize Graph Convolution} & 0.675 & 0.620 \\
    \hline
    \multirow{2}{*}{\tabincell{l}{``NR-ER''\\Tox21, valid subset}} &  \multirow{2}{*}{81/553} & {\scriptsize Logistic Regression} & 0.612 & 0.308 \\\cline{3-5}
    & & {\scriptsize Graph Convolution} & 0.705 & 0.333 \\
    \hline
    \multirow{2}{*}{\tabincell{l}{``HIV\_active''\\HIV, test subset}} &  \multirow{2}{*}{132/4059} & {\scriptsize Logistic Regression} & 0.724 & 0.236 \\\cline{3-5}
    & & {\scriptsize Graph Convolution} & 0.783 & 0.169 \\
    \hline
    \end{tabular}
    \begin{tablenotes}
    \item * Number of positive samples/Number of negative samples
    \end{tablenotes}
\end{table}

To allow better comparison, we propose regression metrics according to previous work on either same models or datasets. For classification datasets, we propose recommended metrics from the two commonly used metrics: AUC-PRC and AUC-ROC. Four representative sets of ROC curves and PRCs are depicted in Figure~\ref{fig:ROCvsPRC}, resulting from the predictions of logistic regression and graph convolutional models on four tasks. Details about these tasks and AUC values of all curves are listed in Table~\ref{tab:ROCvsPRC_AUC}. Note that these four tasks have different class imbalances, represented as the number of positive samples and negative samples.

As noted in previous literature\cite{PRC}, ROC curves and PRCs are highly correlated, but perform significantly differently in case of high class imbalance. As shown in Figure~\ref{fig:ROCvsPRC}, the fraction of positive samples decreases from over 80\% (panels A and B) to less than 5\% (panels G and H). This change accompanies the difference in how the two metrics treat model performances. In particular, PRCs put more emphasis on the low recall (also known as true positive rate (TPR)) side in case of highly imbalanced data: logistic regression slightly outperforms graph convolutional models in the low TPR side of ROC curves (panels C, E and G, lower left corner), which creates different margins on the low recall side of PRCs.

ROC curves and PRCs share one same axis, while using false positive rate (FPR) and precision for the other axis respectively. Recall that FPR and precision are defined as follows:
\begin{equation*}
    \small
    FPR=\frac{\rm{False\,Positive}}{\rm{False\,Positive} + \rm{True\,Negative}}
\end{equation*}
\begin{equation*}
    \small
    Precision=\frac{\rm{True\,Positive}}{\rm{False\,Positive} + \rm{True\,Positive}}
\end{equation*}

When positive samples form only a small proportion of all samples, false positive predictions exert a much greater influence on precision than FPR, amplifying the difference between PRC and ROC curves. Virtual screening experiments do have extremely low positive rates, suggesting that the correct metric to analyze may depend on the experiment at hand. In this work, we hence propose recommended metrics based on positive rates, PRC-AUC is used for datasets with positive rates less than 2\%, otherwise ROC-AUC is used. 

\subsection{Featurization}

A core challenge for molecular machine learning is effectively encoding molecules into fixed-length strings or vectors. Although SMILES strings are unique representations of molecules, most molecular machine learning methods require further information to learn sophisticated electronic or topological features of molecules from limited amounts of data. (Recent work has demonstrated the ability to learn useful representations from SMILES strings using more sophisticated methods\cite{gomez2016automatic}, so it may be feasible to use SMILES strings for further learning tasks in the near future.) Furthermore, the enormity of chemical space often requires representations of molecules specifically suited to the learning task at hand. MoleculeNet contains implementations of six useful molecular featurization methods.

\begin{figure}[h]
  \centering
  \includegraphics[width=.5\textwidth]{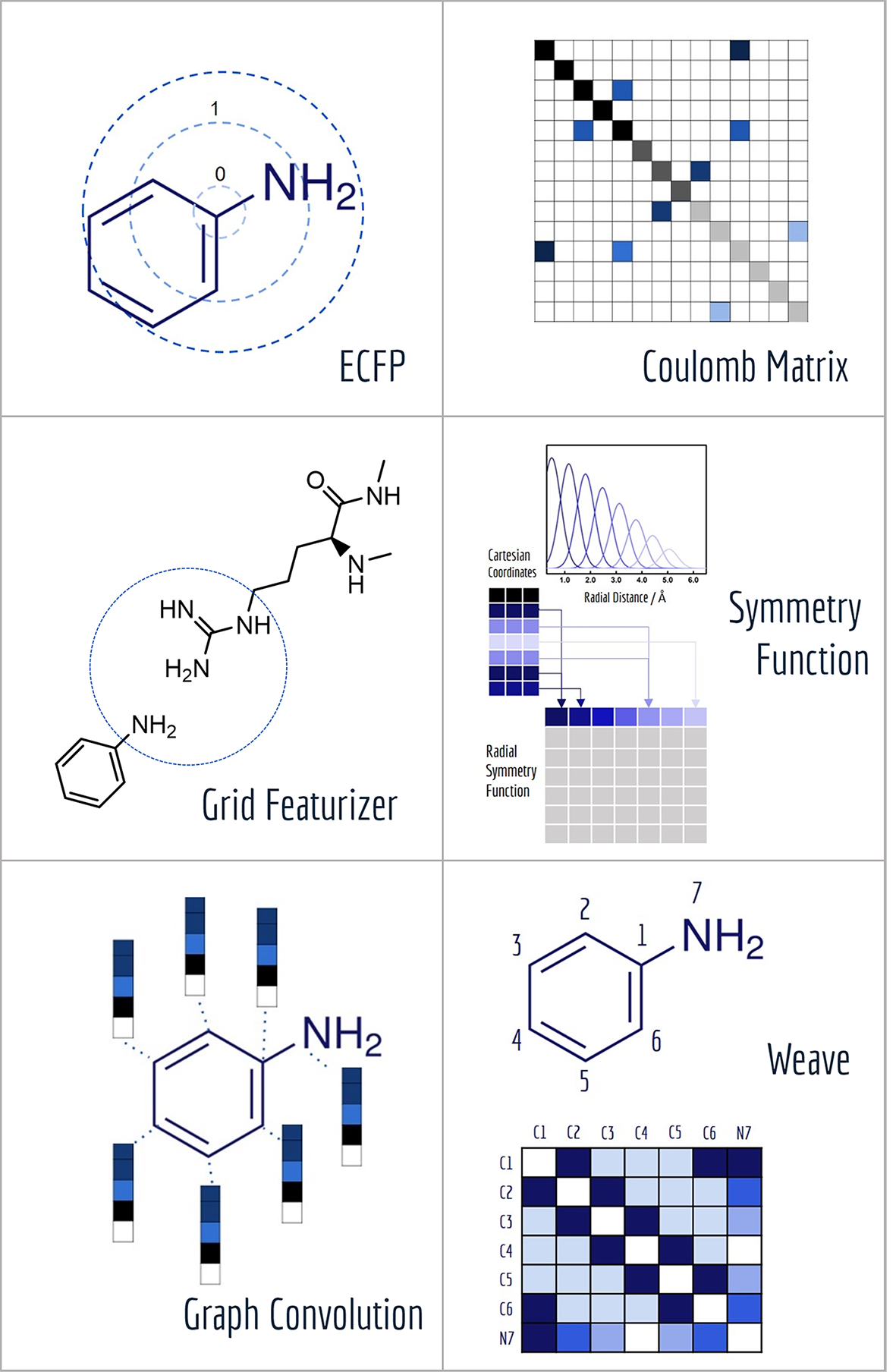}
  \caption{Diagrams of featurizations in MoleculeNet.}
  \label{fig:molnet_feat}
\end{figure}
\subsubsection{ECFP}

Extended-Connectivity Fingerprints (ECFP) are widely-used molecular characterizations in chemical informatics.\cite{ECFP} During the featurization process, a molecule is decomposed into submodules originated from heavy atoms, each assigned with a unique identifier. These segments and identifiers are extended through bonds to generate larger substructures and corresponding identifiers.

After hashing all these substructures into a fixed length binary fingerprint, the representation contains information about topological characteristics of the molecule, which enables it to be applied to tasks such as similarity searching and activity prediction. The MoleculeNet implementation uses ECFP4 fingerprints generated by RDKit.\cite{RDKit}

\subsubsection{Coulomb Matrix}

\textit{Ab-initio} electronic structure calculations typically require a set of nuclear charges \{\textit{Z}\} and the corresponding Cartesian coordinates \{\textbf{R}\} as input.  The Coulomb Matrix (CM) \textbf{M}, proposed by Rupp et al.\cite{GDB7_dataset_prl} and defined below, encodes this information by use of the atomic self-energies and internuclear Coulomb repulsion operator.

\begin{eqnarray*}
    M_{IJ}=
    \begin{cases}
    0.5Z_I^{2.4}\qquad\mbox{for}\;I=J \\
    \frac{Z_IZ_J}{|\bm{R}_I-\bm{R}_J|}\qquad\mbox{for}\;I\neq J
    \end{cases}
\end{eqnarray*}

Here, the off-diagonal elements correspond to the Coulomb repulsion between atoms I and J, and the diagonal elements correspond to a polynomial fit of atomic self-energy to nuclear charge.  The Coulomb Matrix of a molecule is invariant to translation and rotation of that molecule, but not with respect to atom index permutation.  In the construction of coulomb matrix, we first use the nuclear charges and distance matrix generated by RDKit\cite{RDKit} to acquire the original coulomb matrix, then an optional random atom index sorting and binary expansion transformation can be applied during training in order to achieve atom index invariance, as reported by Montavon et al.\cite{GDB7_dataset_arxiv}

\subsubsection{Grid Featurizer}

The grid featurizer is a featurization method (introduced in the current work) initially designed for the PDBbind dataset in which structural information of both the ligand and target protein are considered. Since binding affinity stems largely from the intermolecular forces between ligands and proteins, in addition to intramolecular interactions, we seek to incorporate both the chemical interaction within the binding pocket as well as features of the protein and ligand individually.

The grid featurizer was inspired by the NNscore featurizer \cite{NNscore} and SPLIF \cite{da2014structural} but optimized for speed, robustness, and generalizability. The intermolecular interactions enumerated by the featurizer include salt bridges and hydrogen bonding between protein and ligand, intra-ligand circular fingerprints, intra-protein circular fingerprints, and protein-ligand SPLIF fingerprints. A more detailed breakdown can be found in the Appendix.


\subsubsection{Symmetry Function}

Symmetry function, first introduced by Behler and Parrinello\cite{SymmetryFunction}, is another common encoding of atomic coordinates information. It focuses on preserving the rotational and permutation symmetry of the system. The local environment of an atom in the molecule is expressed as a series of radial and angular symmetry functions  with different distance and angle cutoffs, the former focusing on distances between atom pairs and the latter focusing on angles formed within triplets of atoms. 

As symmetry function put most emphasis on spatial positions of atoms, it is intrinsically hard for it to distinguish different atom types(H, C, O). MoleculeNet utilized a slightly modified version of original symmetry function\cite{ANI-1} which further separate radial and angular symmetry terms according to the type of atoms in the pair or triplet. Further details can be found in the article\cite{ANI-1} or our implementation.

\subsubsection{Graph Convolutions}

The graph convolutions featurization support most graph-based models. It computes an initial feature vector and a neighbor list for each atom. The feature vector summarizes the atom's local chemical environment, including atom-type, hybridization type, and valence structure. Neighbor lists represent connectivity of the whole molecule, which are further processed in each model to generate graph structures (discussed in further details in following parts).

\subsubsection{Weave}

Similar to graph convolutions, the weave featurization encodes both local chemical environment and connectivity of atoms in a molecule. Atomic feature vectors are exactly the same, while connectivity is represented by more detailed pair features instead of neighbor listing. The weave featurization calculates a feature vector for each pair of atoms in the molecule, including bond properties (if directly connected), graph distance and ring info, forming a feature matrix. The method supports graph-based models that utilize properties of both nodes (atoms) and edges (bonds).

\subsection{Models - Conventional Models}

MoleculeNet tests the performance of various machine learning models on the datasets discussed previously. These models could be further categorized into conventional methods and graph-based methods according to their structures and input types. The following sections will give brief introductions to benchmarked algorithms. The results section will discuss performance numbers in detail. Here we briefly review conventional methods including logistic regression, support vector classification, kernel ridge regression, random forests\cite{breiman2001random}, gradient boosting\cite{friedman2001greedy}, multitask networks\cite{ramsundar2015massively, ma2015deep}, bypass networks \cite{Bypass_network} and influence relevance voting\cite{IRV}. The next section graph-based models will give introductions to graph convolutional models\cite{graphconv_feat}, weave models\cite{kearnes2016graphconv}, directed acyclic graph models\cite{lusci2013deep}, deep tensor neural networks\cite{schutt2016quantum}, ANI-1\cite{ANI-1} and message passing neural networks\cite{MPNN}. As part of this work, all methods are implemented in the open source DeepChem package\cite{deepchem}.

\subsubsection{Logistic Regression}

Logistic regression models (Logreg) apply the logistic function to weighted linear combinations of their input features to obtain model predictions. It is often common to use regularization to encourage learned weights to be sparse. \cite{friedman2000additive} Note that logistic regression models are only defined for classification tasks.

\subsubsection{Support Vector Classification}

Support vector machine (SVM) is one of the most famous and widely-used machine learning method.\cite{cortes1995support} As in classification task, it defines a decision plane which separates data points of different class with maximized margin. To further increase performance, we incorporates regularization and a radial basis function kernel (KernelSVM).

\subsubsection{Kernel Ridge Regression}

Kernel ridge regression(KRR) is a combination of ridge regression and kernel trick. By using a nonlinear kernel function(radial basis function), it learns a non-linear function in the original space that maps features to predicted values.

\subsubsection{Random Forests}

Random forests (RF) are ensemble prediction methods.\cite{breiman2001random} A random forest consists of many individual decision trees, each of which is trained on a subsampled version of the original dataset. The results for individual trees are averaged to provide output predictions for the full forest. Random forests can be used for both classification and regression tasks. Training a random forest can be computationally intensive, so benchmarks only include random forest results for smaller datasets.

\subsubsection{Gradient Boosting}

Gradient boosting is another ensemble method consisting of individual decision trees.\cite{friedman2001greedy} In contrast to random forests, it builds relatively simple trees which are sequentially incorporated to the ensemble. In each step, a new tree is generated in a greedy manner to minimize loss function. A sequence of such "weak" trees are combined together into an additive model. We utilize the XGBoost implementation of gradient boosting in DeepChem.\cite{XGBoost}

\subsubsection{Multitask/Singletask Network}

In a multitask network,\cite{ramsundar2015massively} input featurizations are processed by fully connected neural network layers. The processed output is shared among all learning tasks in a dataset, and then fed into separate linear classifiers/regressors for each different task. In the case that a dataset contains only a single task, multitask networks are just fully connected neural networks(Singletask Network). Since multitask networks are trained on the joint data available for various tasks, the parameters of the shared layers are encouraged to produce a joint representation which can share information between learning tasks. This effect does seem to have limitations; merging data from uncorrelated tasks has only moderate effect.\cite{kearnes2016modeling} As a result, MoleculeNet does not attempt to train extremely large multitask networks combining all data for all datasets.

\subsubsection{Bypass Multitask Networks}

Multitask modeling relies on the fact that some features have explanatory power that is shared among multiple tasks. Note that the opposite may well be true; features useful for one task can be detrimental to other tasks. As a result, vanilla multitask networks can lack the power to explain unrelated variations in the samples. Bypass networks attempt to overcome this variation by merging in per-task independent layers that ``bypass'' shared layers to directly connect inputs with outputs.\cite{Bypass_network} In other words, bypass multitask networks consist of $n_\text{tasks}+1$ independent components: one ``multitask'' layer mapping all inputs to shared representations, and $n_\text{tasks}$ ``bypass'' layers mapping inputs for each specific task to their labels. As the two groups have separate parameters, bypass networks may have greater explanatory power than vanilla multitask networks.

\subsubsection{Influence Relevance Voting}

Influence Relevance Voting (IRV) systems are refined K-nearest neighbor classifiers.\cite{IRV} Using the hypothesis that compounds with similar substructures have similar functionality, the IRV classifier makes its prediction by combining labels from the top K compounds most similar to a provided test sample.

The Jaccard-Tanimoto similarity between fingerprints of compounds is used as the similarity measurement:

\begin{equation*}
    S(\vec{A},\vec{B})=
    \frac{A\cap B}{A\cup B}
\end{equation*}

Then IRV model calculates a weighted sum of the labels of top K similar compounds to predict the result, in which weights are the outputs of a one-hidden layer neural network with similarities and rankings of top K compounds as input. Detailed descriptions of the model can be found in the original article.\cite{IRV} 

\subsection{Models - Graph Based Models}

\begin{figure}[h]
  \centering
  \includegraphics[width=\textwidth]{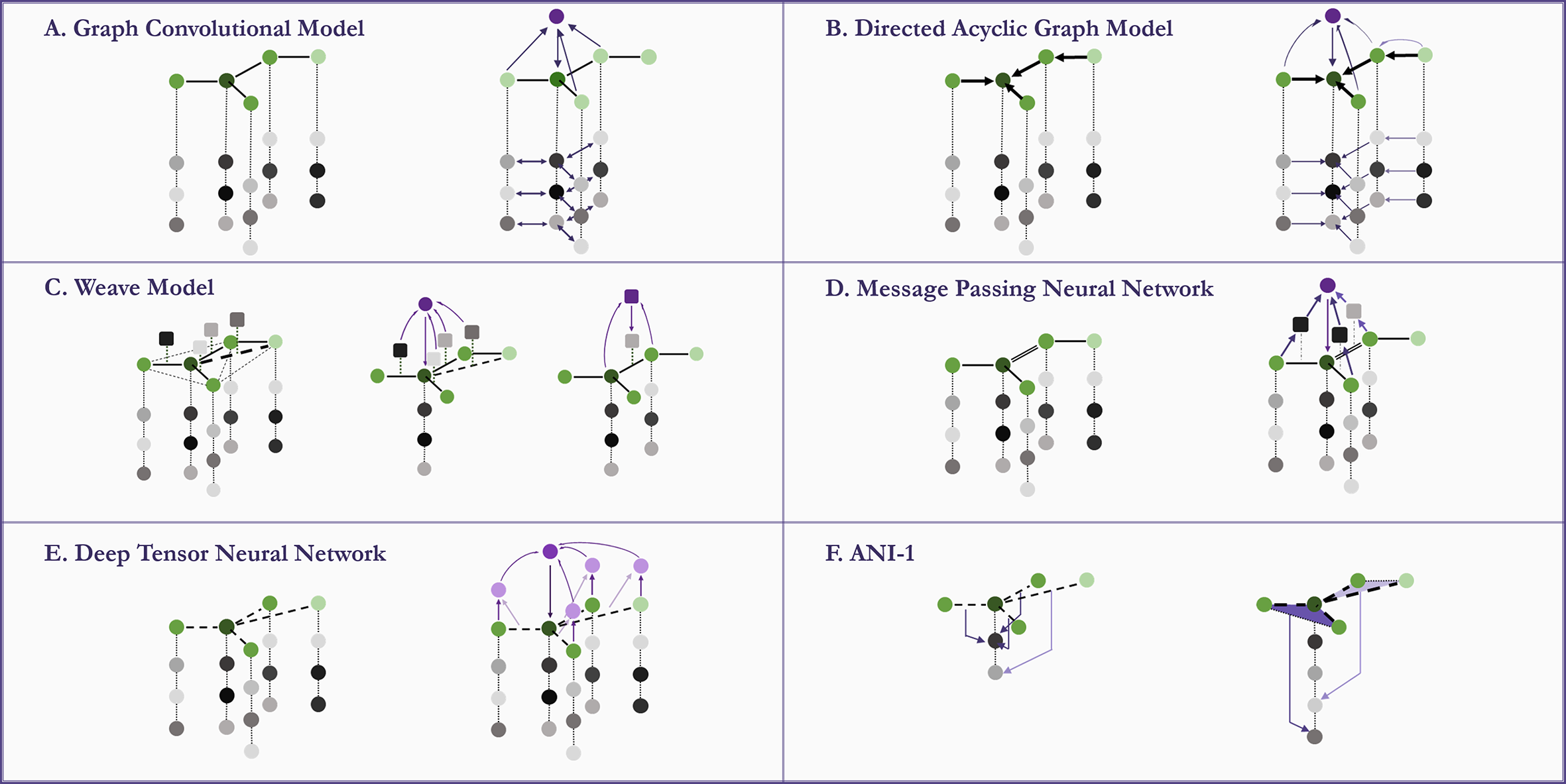}
  \caption{Core structures of graph-based models implemented in MoleculeNet. To build features for the central dark green atom: \textbf{A} Graph Convolutional Model: features are updated by combination with neighbor atoms; \textbf{B} Directed Acyclic Graph Model: all bonds are directed towards the central atom, features are propagated from the farthest atom to the central atom through directed bonds; \textbf{C} Weave Model: Pairs are formed between each pair of atoms(including not directly bonded pairs), features for the central atom are updated using all other atoms and their corresponding pairs, pair features are also updated by combination of the two pairing atoms; \textbf{D} Message Passing Neural Network: Neighbor atoms' features are input into bond-type dependent neural networks, forming outputs(messages). Features of the central atom are then updated using the outputs; \textbf{E} Deep Tensor Neural Network: No explicit bonding information is included, features are updated using all other atoms based on their corresponding physical distances; \textbf{F} ANI-1: features are built on distance information between pairs of atoms(radial symmetry functions) and angular information between triplets of atoms(angular symmetry functions).}
  \label{fig:model_structure1}
\end{figure}

Early attempts to directly use molecular structures instead of selected features has emerged in 1990s\cite{baskin1997neural, kireev1995chemnet}. While in recent years, models propelled by the very similar idea start to grow rapidly. These specifically designed methods, namely graph-based models, are naturally suitable for modeling molecules. By defining atoms as nodes, bonds as edges, molecules can be modeled as mathematical graphs. As noted in a recent paper\cite{MPNN}, this natural similarity has inspired a number of models to utilize the graph structure of molecules to gain higher performances. In general, graph-based models apply adaptive functions to nodes and edges, allowing for a learnable featurization process. MoleculeNet provides implementations of multiple graph-based models which use different variants of molecular graphs. We describe these methods in the following sections. Figure~\ref{fig:model_structure1} provide simple illustrations of these methods' core structures.

\subsubsection{Graph Convolutional models}

Graph convolutional models (GC) extend the decomposition principles of circular fingerprints. Both methods gradually merge information from distant atoms by extending radially through bonds. This information is used to generate identifiers for all substructures. However, instead of applying fixed hash functions, graph convolutional models allow for adaptive learning by using differentiable network layers. This creates a learnable process capable of extracting useful representations of molecules suited to the task at hand. (Note that this property is shared, to some degree, by all deep architectures considered in MoleculeNet. However, graph convolutional architectures are more explicitly designed to encourage extraction of useful featurizations).

On a higher level, graph convolutional models treat molecules as undirected graphs, and apply the same learnable function to every node (atom) and its neighbors (bonded atoms) in the graph. This structure recapitulates convolution layers in visual recognition deep networks.

MoleculeNet uses the graph convolutional implementation in DeepChem from previous work.\cite{altae2016low} This implementation converts SMILES strings into molecular graphs using RDKit \cite{RDKit} As mentioned previously, the initial representations assign to each atom a vector of features including its element, connectivity, valence, etc. Then several graph convolutional modules, each consisting of a graph convolutional layer, a batch normalization layer and a graph pool layer, are sequentially added, followed by a fully-connected dense layer. Finally, the feature vectors for all nodes (atoms) are summed, generating a graph feature vector, which is fed to a classification or regression layer.

\subsubsection{Weave models}

The Weave architecture is another graph-based model that regards each molecule as a undirected graph. Similar to graph convolutional models, it utilizes the idea of adaptive learning on extracting meaningful representations.\cite{kearnes2016graphconv} The major difference is the size of the convolutions: To update features of an atom, weave models combine information from all other atoms and their corresponding pairs in the molecule. Weave models are more efficient at transmitting information between distant atoms, at the price of increased complexity for each convolution.

In our implementation, a molecule is first encoded into a list of atomic features and a matrix of pair features by the weave model's featurization method. Then in each weave module, these features are input into four sets of fully connected layers (corresponding to four paths from two original features to two updated features) and concatenated to form new atomic and pair features. After stacking several weave modules, a similar gather layer combines atomic features together to form molecular features that are fed into task-specific layers.

\subsubsection{Directed Acyclic Graph models}

Directed Acyclic Graph (DAG) models regard molecules as directed graphs. While chemical bonds typically do not have natural directions, one can arbitrarily generate a DAG on a molecule by designating a central atom and then define directions of all bonds in certain orientations towards the atom.\cite{lusci2013deep} In the case of small molecules, taking all possible orientations is computationally feasible. In other words, for a molecule with $n_a$ atoms, the model will generate $n_a$ DAGs, each centered on a different atom. 

In the actual calculations of a graph, a vector of graph features is calculated for each atom based on its atomic features (reusing the graph convolutions featurizer) and its parents' graph features. As features gradually propagate through bonds, information converges on the central atom. Then a final sum of all graphs gives the molecular features, which are fed into classification or regression tasks. Note that $n_a$ graphs are evaluated for each molecule, which can cause a significant increase in required calculations.

\subsubsection{Deep Tensor Neural Networks}

Deep Tensor Neural Networks (DTNN) are adaptable extensions of the Coulomb Matrix featurizer.\cite{schutt2016quantum} The core idea is to directly use nuclear charge (atom number) and the distance matrix to predict energetic, electronic or thermodynamic properties of small molecules. To build a learnable system, the model first maps atom numbers to trainable embeddings(randomly initialized) as atomic features. Then each atomic feature $a_i$ is updated based on distance information $d_{ij}$ and other atomic features $a_j$. Comparing with Weave models, DTNNs share the same idea in terms of updating based on both atomic and pair features, while the difference is using physical distance instead of graph distance. Note that the use of 3D coordinates to calculate physical distances limits DTNNs to quantum mechanical (or perhaps biophysical) datasets.

We reimplement the model proposed by Sch\"{u}tt et al.\cite{schutt2016quantum} in a more generalized fashion. Atom numbers and a distance matrix are calculated by RDKit\cite{RDKit}, using the Coulomb matrix featurizer. After embedding atom numbers into feature vectors $a_i$, we update $a_i$ in each convolutional layer by adding the outputs from all network layers which use $d_{ij}$ and $a_j$ ($i\neq j$) as input. After several layers of convolutions, all atomic features are summed together to form molecular features, used for classification and regression tasks.

\subsubsection{ANI-1}

ANI-1 is designed as a deep neural network capable of learning accurate and transferable potentials for organic molecules. It is based on the symmetry function method\cite{SymmetryFunction}, with additional changes enabling it to learn different potentials for different atom types. Feature vector, a series of symmetry functions, is built for each atom in the molecule based on its atom type and interaction with other atoms. Then the feature vectors are fed into different neural network potentials(depending on atom types) to generate predictions of properties.

This model is first introduced by Smith et al.\cite{ANI-1}. In their original article, the model is trained on ~58k small molecules with 8 or less heavy atoms, each with multiple poses and potentials. Training set in total has ~17.2 million data points, which is far bigger than qm8 or qm9 in our collection. Since we only have molecules in their most stable configuration, we cannot expect similar level of accuracy. Further comparison and benchmarking with similar size of training set is left to future work.

\subsubsection{Message Passing Neural Networks}

Message Passing Neural Network(MPNN) is a generalized model proposed by Gilmer et al.\cite{MPNN} that targets to formulate a single framework for graph based model. The prediction process is separated into two phases: message passing phase and readout phase. Multiple message passing phases are stacked to extract abstract information of the graph, then the readout phase is responsible for mapping the graph to its properties.

Here we reimplemented the best-performing model in the original article: using an edge network as message passing function and a set2set model\cite{set2set} as readout function. In message passing phase, an edge-dependent neural network maps all neighbor atoms' feature vectors to updated messages, which are then merged using gated recurrent units. In the final readout phase, feature vectors for all atoms are regarded as a set, then an LSTM with attention mechanism is applied on the top for multiple steps, exporting the final state as outputs for the molecule.

\section{Results and Discussion}

In this section, we discuss the performances of benchmarked models on MoleculeNet datasets. Different models are applied depending on the size, features and task types of the dataset. All graph models use their corresponding featurizations. Non-graph models use ECFP featurizations by default, Coulomb Matrix (CM) and Grid featurizer are also applied for certain datasets.

We run a brief Gaussian process hyperparameter optimization on each combination of dataset and model. Then three independent runs with different random seeds are performed. More detailed description of optimization method and performance tables can be found in the Appendix. Note that all benchmark results presented here are the average of three runs, with standard deviations listed or illustrated as error bars.

We also run a set of experiments focusing on how variable size of training set affect model performances.(Tox21, FreeSolv and QM7) Details will be presented in the following texts.

\subsection{Physiology and Biophysics Tasks}

\begin{figure}[htbp]
  \centering
  \includegraphics[width=\textwidth]{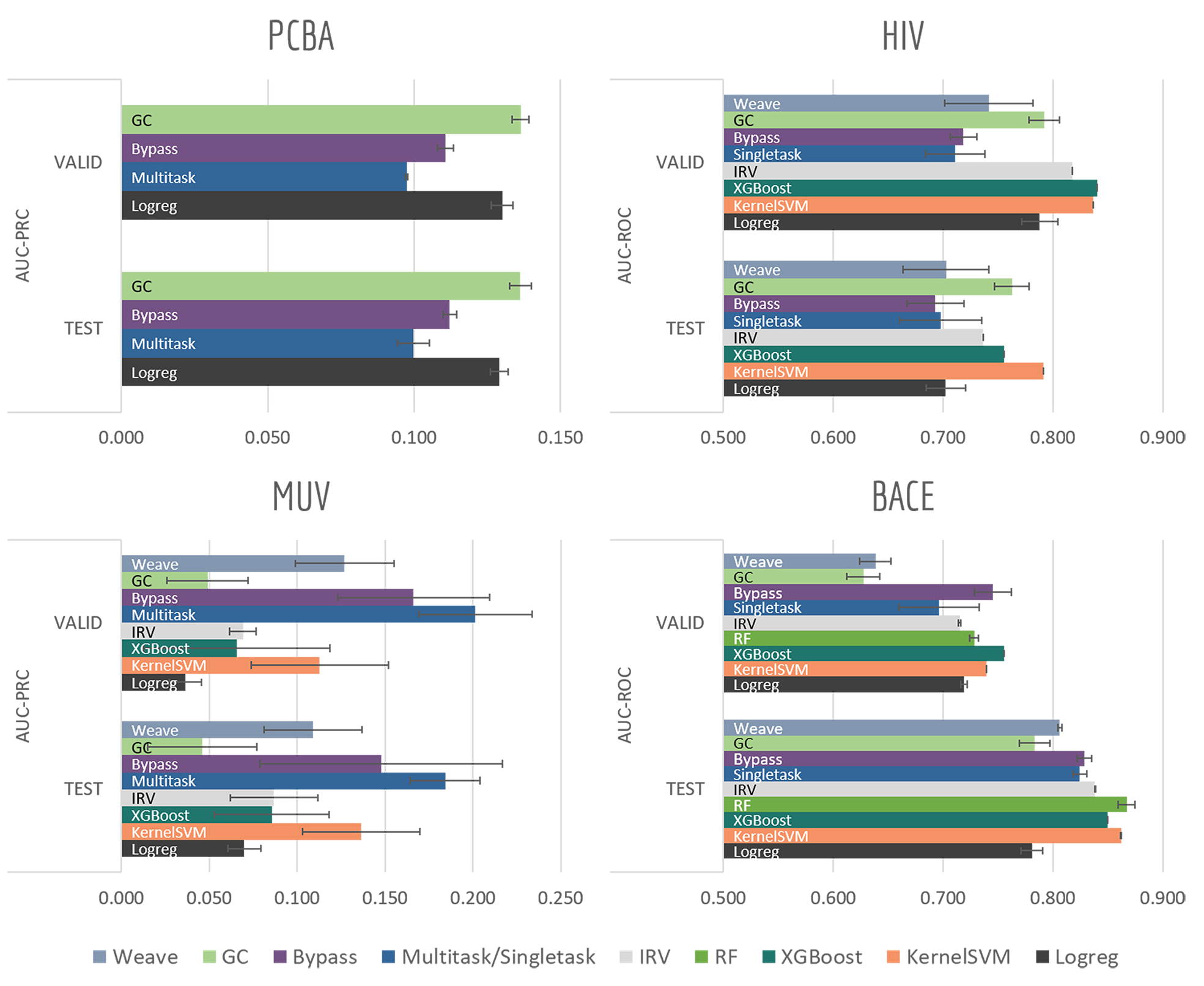}
  \caption{Benchmark performances for biophysics tasks: \textbf{PCBA}, 4 models are evaluated by AUC-PRC on random split; \textbf{MUV}, 8 models are evaluated by AUC-PRC on random split; \textbf{HIV}, 8 models are evaluated by AUC-ROC on scaffold split; \textbf{BACE}, 9 models are evaluated by AUC-ROC on scaffold split. For AUC-ROC and AUC-PRC, higher value indicates better performance(to the right).}
  \label{fig:PCBA_HIV_MUV_BACE}
\end{figure}

\begin{figure}[htbp]
  \centering
  \includegraphics[width=\textwidth]{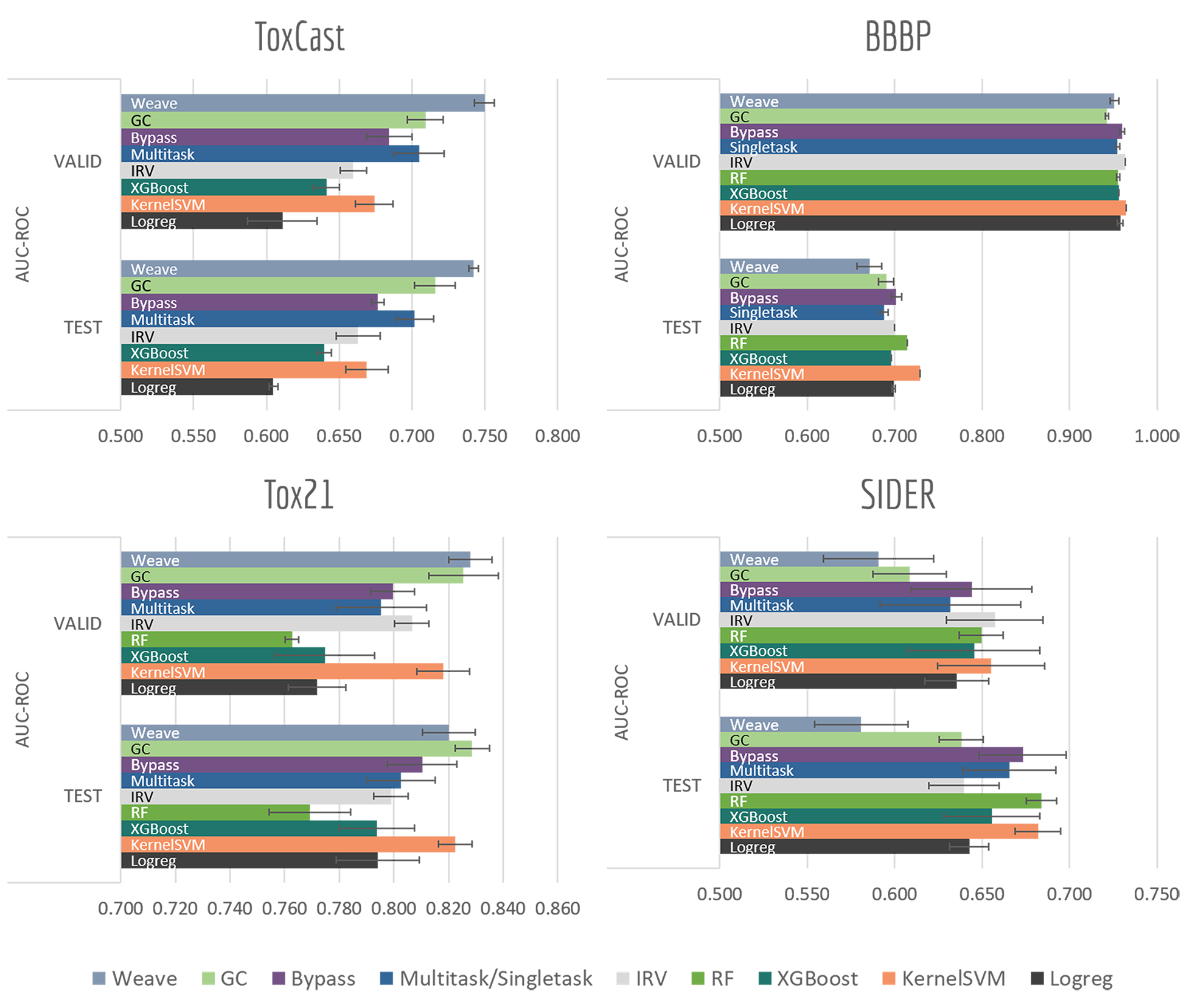}
  \caption{Benchmark performances for physiology tasks: \textbf{ToxCast}, 8 models are evaluated by AUC-ROC on random split; \textbf{Tox21}, 9 models are evaluated by AUC-ROC on random split; \textbf{BBBP}, 9 models are evaluated by AUC-ROC on scaffold split; \textbf{SIDER}, 9 models are evaluated by AUC-ROC on random split. For AUC-ROC, higher value indicates better performance(to the right).}
  \label{fig:ToxCast_BBBP_Tox21_SIDER}
\end{figure}

\begin{figure}[htbp]
  \centering
  \includegraphics[width=.5\textwidth]{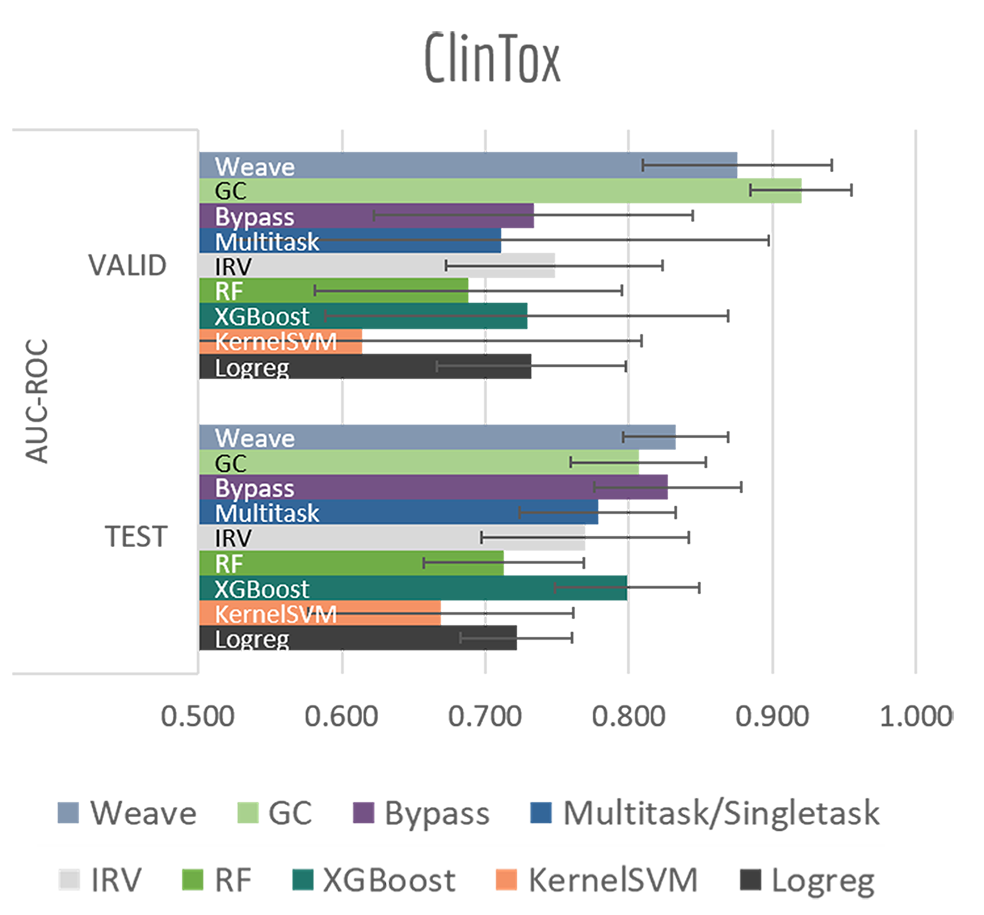}
  \caption{Benchmark performances for physiology tasks: \textbf{ClinTox}, 9 models are evaluated by AUC-ROC on random split.}
  \label{fig:ClinTox}
\end{figure}

Tables~\ref{tab:PCBA_MUV_HIV_BACE}, \ref{tab:BBBP_Tox21_ToxCast_SIDER} and Figures~\ref{fig:PCBA_HIV_MUV_BACE}, \ref{fig:ToxCast_BBBP_Tox21_SIDER}, \ref{fig:ClinTox} report AUC-ROC or AUC-PRC results of $4$ to $9$ different models on biophysics datasets (PCBA, MUV, HIV, BACE) and physiology datasets (BBBP, Tox21, Toxcast, SIDER, ClinTox). Some models were too computationally expensive to be run on the larger datasets. All of these datasets contain only classification tasks. 

Most models have train scores (listed in Tables~\ref{tab:PCBA_MUV_HIV_BACE}, \ref{tab:BBBP_Tox21_ToxCast_SIDER}) higher than validation/test scores, indicating that overfitting is a general issue. Singletask logistic regression exhibits the largest gaps between train scores and validation/test scores, while models incorporating multitask structure generally show less overfit, suggesting that multitask training has a regularizing effect. Most physiological and biophysical datasets in MoleculeNet have only a low volume of data for each task. Multitask algorithms combine different tasks, resulting in a larger pool of data for model training. In particular, multitask training can, to some extent, compensate for the limited data amount available for each individual task.

Graph convolutional models and weave models, each based on an adaptive method of featurization\cite{graphconv_feat, kearnes2016graphconv}, show strong validation/test results on larger datasets, along with less overfit. Similar results are reported in previous graph-based algorithms \cite{kearnes2016graphconv, graphconv_feat, MPNN, schutt2016quantum, lusci2013deep}, showing that learnable featurizations can provide a large boost compared with conventional featurizations.

For smaller singletask datasets (less than 3000 samples), differences between models are less clear. Kernel SVM and ensemble tree methods (gradient boosting and random forests) are more robust under data scarcity, while they generally need longer running time (see Table~\ref{tab:running_time}). Worse performances of graph-based models are within expectation as complex models generally require more training data.

Bypass networks show higher train scores and equal or higher validation/test scores compared with vanilla multitask networks, suggesting that the bypass structure does add robustness. IRV models achieve performance broadly comparable with multitask networks. However, the quadratic nearest neighbor search makes the IRV models slower to train than the multitask networks (see Table~\ref{tab:running_time}).

Three datasets (HIV, BACE, BBBP) in these two categories are evaluated under scaffold splitting. As compounds are divided by their molecular scaffolds, increasing differences between train, validation and test performances are observed. Scaffold splits provide a stronger test of a given model's generalizability compared with random splitting. Two datasets (PCBA, MUV) are evaluated by AUC-PRC, which is more practically useful under high class imbalance as discussed above. Graph convolutional model performs the best on PCBA (positive rate $1.40\%$), while results on MUV (positive rate $0.20\%$) are much less stable, which is most likely due to its extreme low amount of positive samples. Under such high imbalance, graph-based models are still not robust enough in controlling false positives.

\begin{figure}[htbp]
  \centering
  \includegraphics[width=.8\textwidth]{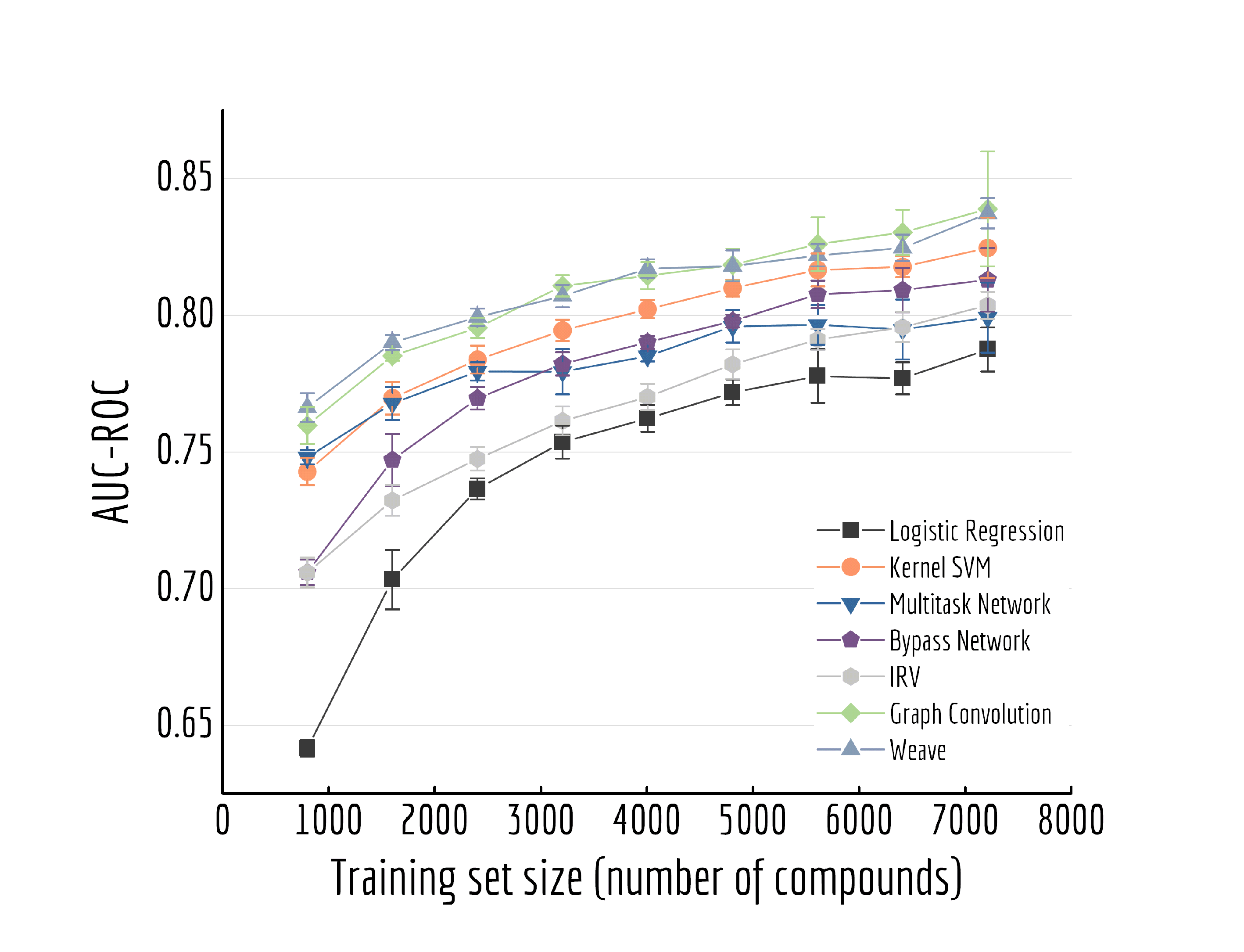}
  \caption{Out-of-sample performances with different training set sizes on Tox21. Each datapoint is the average of 5 independent runs, with standard deviations shown as error bars.}
  \label{fig:Tox21_variable}
\end{figure}

Here we performed a more detailed experiment to illustrate how model performances change with increasing training samples. We trained multiple models on Tox21 with training sets of different size($10\%$ to $90\%$ of the whole dataset) Figure~\ref{fig:Tox21_variable} displayed mean out-of-sample performances (and standard deviations) of five independent runs. A clear increase on performance is observed for each model, and graph-based models (Graph convolutional model and weave model) always stay on top of the lines. By drawing a horizontal line at around 0.80, we can see graph-based models achieve the similar level of accuracy with multitask networks by using only one-third of the training samples($30\%$ versus $90\%$).

\subsection{Biophysics Task - PDBbind}

\begin{figure}[htbp]
  \centering
  \includegraphics[width=0.5\textwidth]{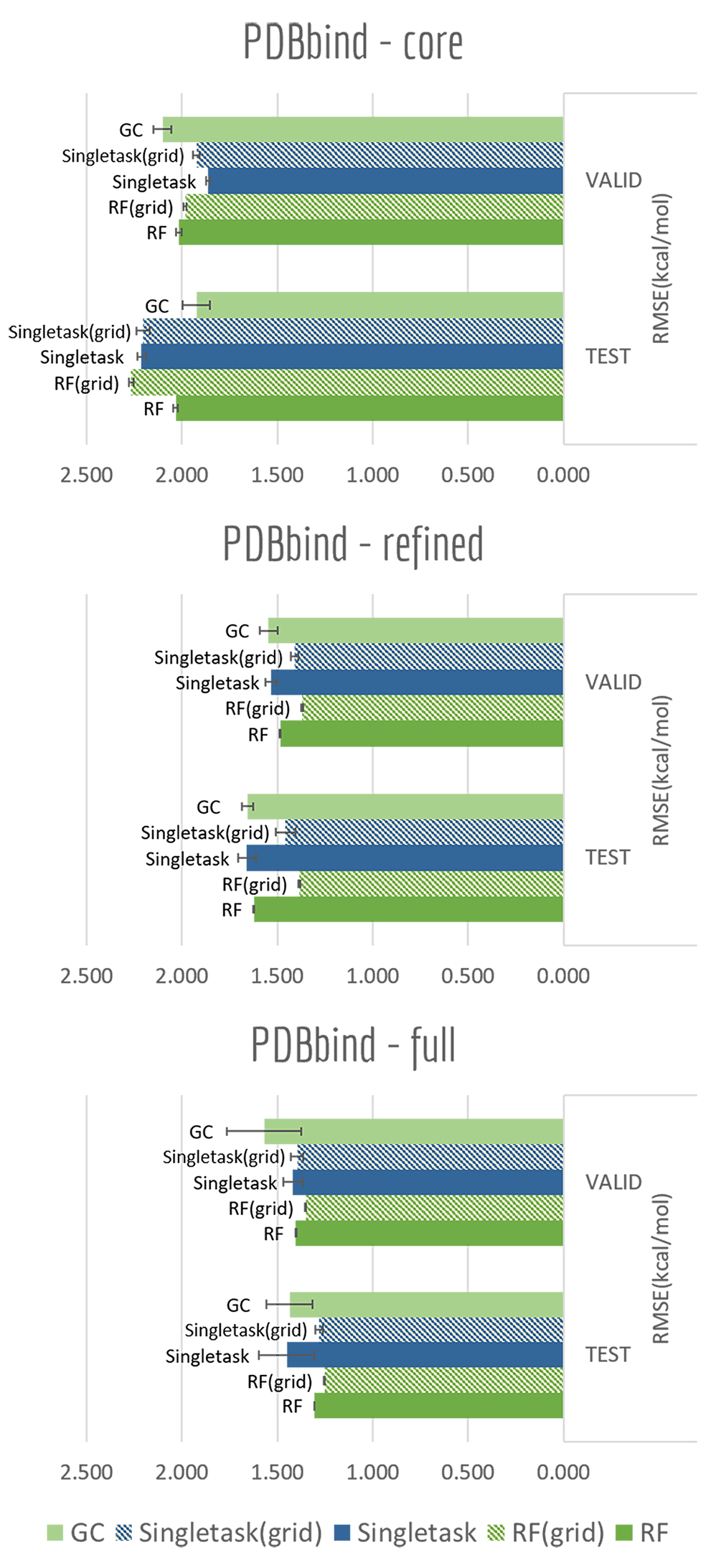}
  \caption{Benchmark performances of \textbf{PDBbind}: 5 models are evaluated by RMSE on the three subsets: core, refined and full. Time split is applied to all three subsets. Noe that for RMSE, lower value indicates better performance(to the right).}
  \label{fig:PDBbind}
\end{figure}

The PDBBind dataset maps distinct ligand-protein structures to their binding affinities. As discussed in the datasets section, we created grid featurizer to harness the joint ligand-protein structural information in PDBBind to build a model that predicts the experimental $K_i$ of binding. We applied time splitting to all three subsets: core, refined, and full subsets of PDBbind(Core contains roughly 200 structures, refined 4000, and full 15000. The smaller datasets are cleaned more thoroughly than larger datasets.), with all results displayed in Table~\ref{tab:PDBbind} and Figure~\ref{fig:PDBbind}. Clearly as dataset size increased, we can see a significant boost on validation/test set performances. At the same time, for the two larger subsets: refined and full, switching from pure ligand-based ECFP to grid featurizer do increase the performances by a small margin in both Singletask networks and random forests. While for core subset, all models are showing relatively high errors and two featurizations do not show clear differences, which is within expectation as sample amount in core subset is too small to support a stable model performance. Note that models on the full set aren't significantly superior to models with less data; this effect may be due to the additional data being less clean.

Note that all models display heavy overfitting. Additional clean data may be required to create more accurate models for protein-ligand binding.

\subsection{Physical Chemistry Tasks}

\begin{figure}[htbp]
  \centering
  \includegraphics[width=.5\textwidth]{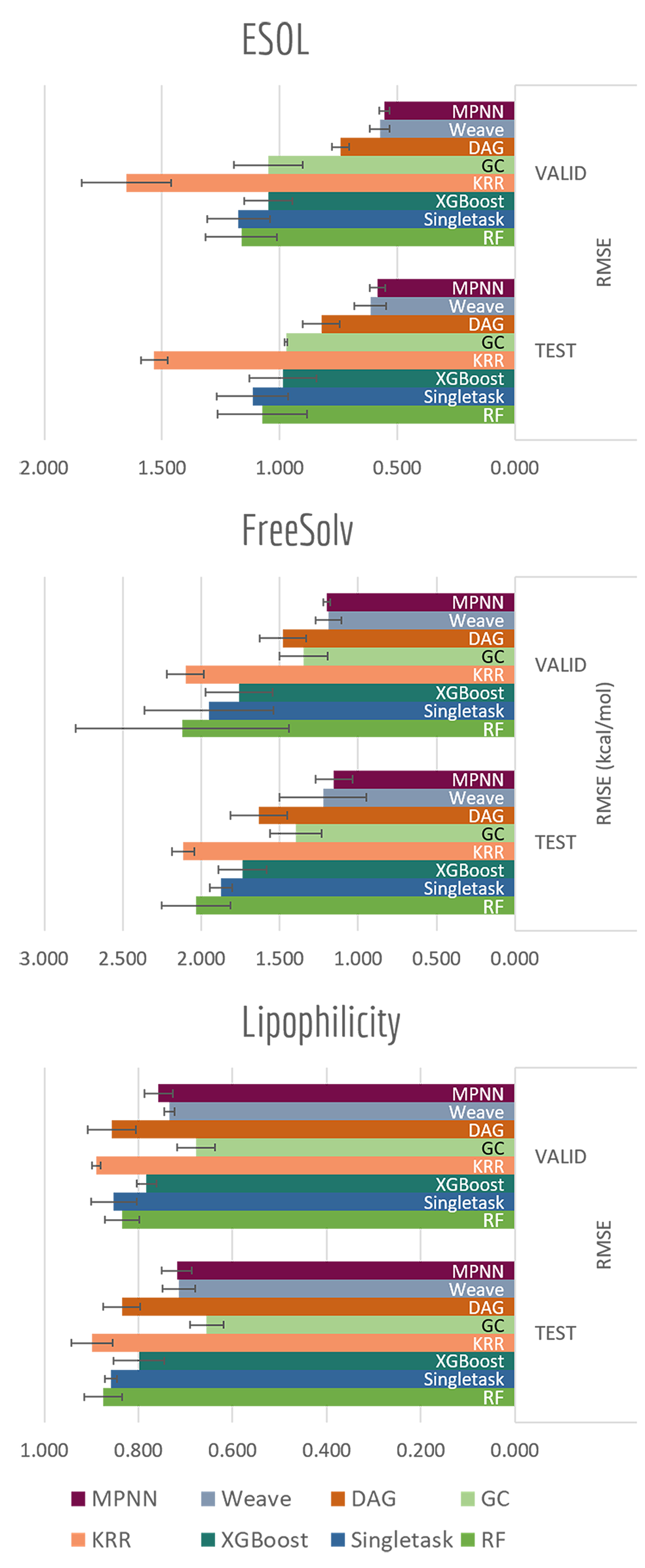}
  \caption{Benchmark performances for physical chemistry tasks: \textbf{ESOL}, 8 models are evaluated by RMSE on random split; \textbf{FreeSolv}, 8 models are evaluated by RMSE on random split; \textbf{Lipophilicity}, 8 models are evaluated by RMSE on random split. Note that for RMSE, lower value indicates better performance(to the right).}
  \label{fig:ESOL_FreeSolv_Lipo}
\end{figure}

Solubility, solvation free energy and lipophilicity are basic physical chemistry properties important for understanding how molecules interact with solvents. Figure~\ref{fig:ESOL_FreeSolv_Lipo} and Table~\ref{tab:ESOL_FreeSolv_Lipo} presented performances on predicting these properties. 

Graph-based methods: graph convolutional model, DAG, MPNN and weave model all exhibit significant boosts over vanilla singletask network, indicating the advantages of learnable featurizations. Differences between graph-based methods are rather minor and task-specific. The best-performing models in this category can already reach the accuracy level of \textit{ab-initio} predictions(+/- 0.5 for ESOL, +/- 1.5 kcal/mol for FreeSolv).

We performed a more detailed comparison between data-driven methods and \textit{ab-initio} calculations on FreeSolv. Hydration free energy has been widely used as a test of computational chemistry methods. With free energy values ranging from $-25.5$ to $3.4$ $\mbox{kcal/mol}$ in the FreeSolv dataset, RMSE for calculated results reached up to $1.5$ $\mbox{kcal/mol}$.\cite{SAMPL4} On the other hand, though machine learning methods typically need large amounts of training data to acquire predictive power, they can achieve higher accuracies given enough data. We investigated how the performance of machine learning methods on FreeSolv changes with the volume of training data. In particular, we want to know the amount of data required for machine learning to achieve accuracy similar to that of physically inspired algorithms.

\begin{figure}[htbp]
  \centering
  \includegraphics[width=.8\textwidth]{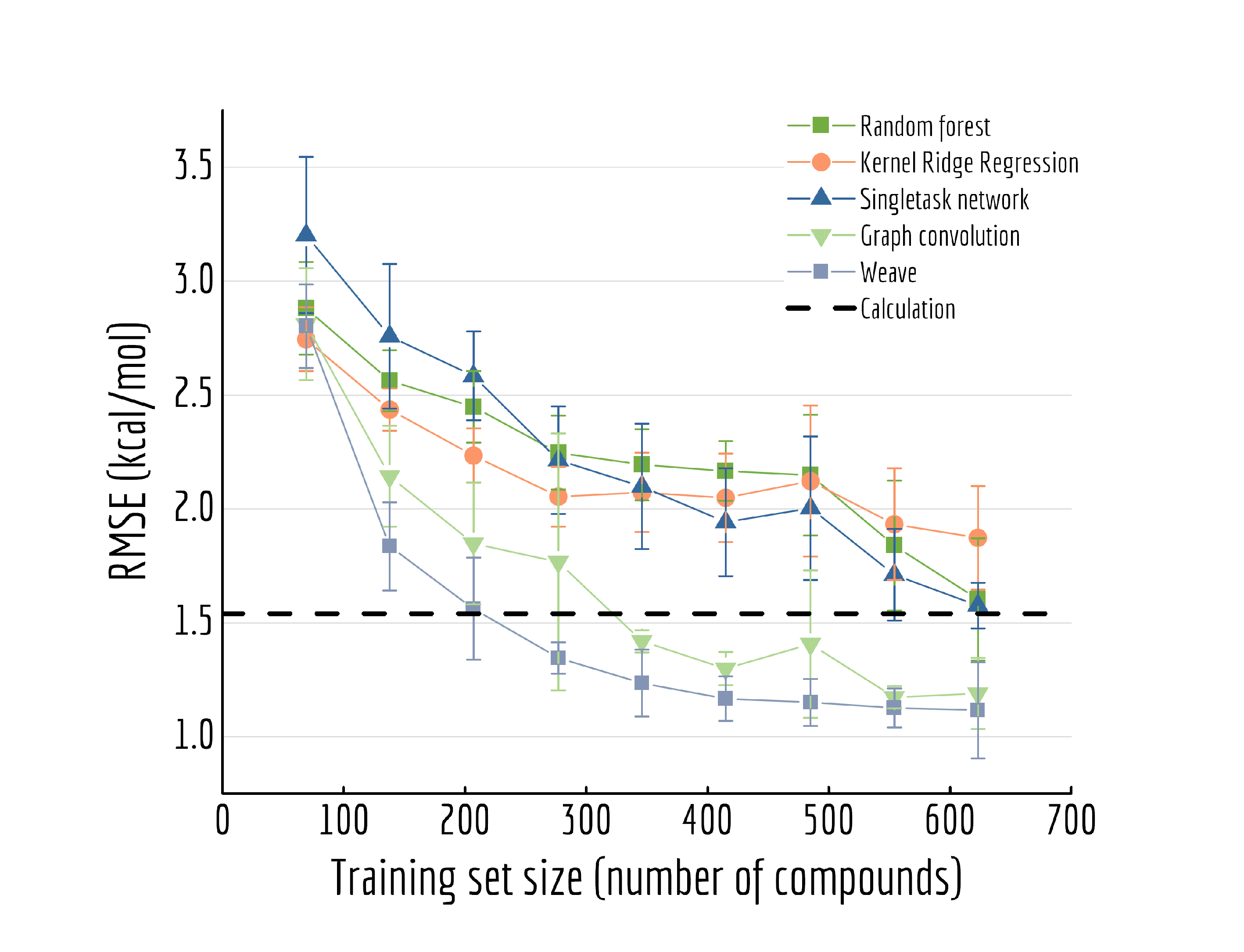}
  \caption{Out-of-sample performances with different training set sizes on FreeSolv. Each datapoint is the average of 5 independent runs, with standard deviations shown as error bars.}
  \label{fig:FreeSolv_variable}
\end{figure}

For Figure~\ref{fig:FreeSolv_variable}, we similarly generated a series of models with different training set volumes and calculated their out-of-sample RMSE. Each data point displayed is the average of 5 independent runs, with standard deviations displayed as error bars. Both graph convolutional model and weave model are capable of achieving better performances with enough training samples ($50\%$ and $30\%$ of the data respectively). Given the size of FreeSolv dataset is only around 600 compounds, a weave model can reach state-of-the-art free energy calculation performances by training on merely 200 samples. On the other hand, comparing with singletask network's performance, weave model achieved the same level of accuracy with only one-third of the training samples.

\subsection{Quantum Mechanics Tasks}

\begin{figure}[htbp]
  \centering
  \includegraphics[width=.43\textwidth]{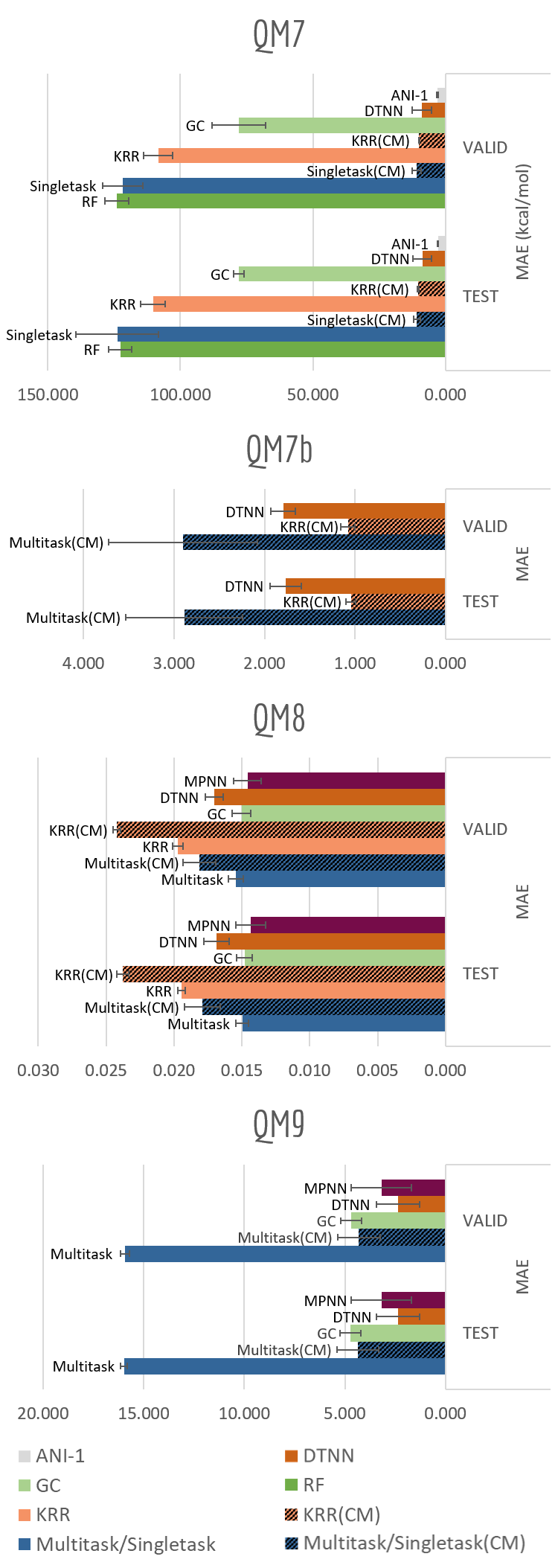}
  \caption{Benchmark performances for quantum mechanics tasks: \textbf{QM7}, 8 models are evaluated by MAE on stratified split; \textbf{QM7b}, 3 models (QM7b only provides 3D coordinates) are evaluated by MAE on random split; \textbf{QM8}, 7 models are evaluated by MAE on random split; \textbf{QM9}, 5 models are evaluated by MAE on random split. Note that for MAE, lower value indicates better performance(to the right)}
  \label{fig:QM}
\end{figure}

The QM datasets (QM7, QM7b, QM8, QM9) represent another distinct category of properties that are typically calculated through solving Schr\"odinger's equation (approximately using techniques such as DFT). As most conventional methods are slower than data-driven methods by orders of magnitude, we hope to learn effective approximators by training on existing datasets.

Table~\ref{tab:QM7_QM7b_QM8_QM9} and Figure~\ref{fig:QM} display the performances in mean absolute error of multiple methods. Table~\ref{tab:QM7b}, \ref{tab:QM8} and \ref{tab:QM9} show detailed performances for each task.(Due to difference in range of labels, mean performances of QM7b and QM9 are more skewed) Unsurprisingly, significant boosts on performances and less overfitting are observed for models incorporating distance information (multitask networks and KRR with Coulomb Matrix featurization, ANI-1, DTNN, MPNN). In particular, KRR and multitask networks(CM) outperform their corresponding baseline models in QM7 and QM9 by a large margin, while ANI-1, DTNN and MPNN display less error comparing with graph convolutional models as well. At the same time, graph-based methods gain better performances than multitask networks and KRR (CM) on most tasks. Table~\ref{tab:QM7b} shows that DTNN outperforms KRR(CM) on 12/14 tasks in QM7b(Though the mean error shows the opposite result due to averaging errors on different magnitudes). In total, ANI-1, DTNN and MPNN covered the best-performing models on 28/39 of all tasks in this category, again reflecting the superiority of learnable featurization.

\begin{figure}[htbp]
  \centering
  \includegraphics[width=.8\textwidth]{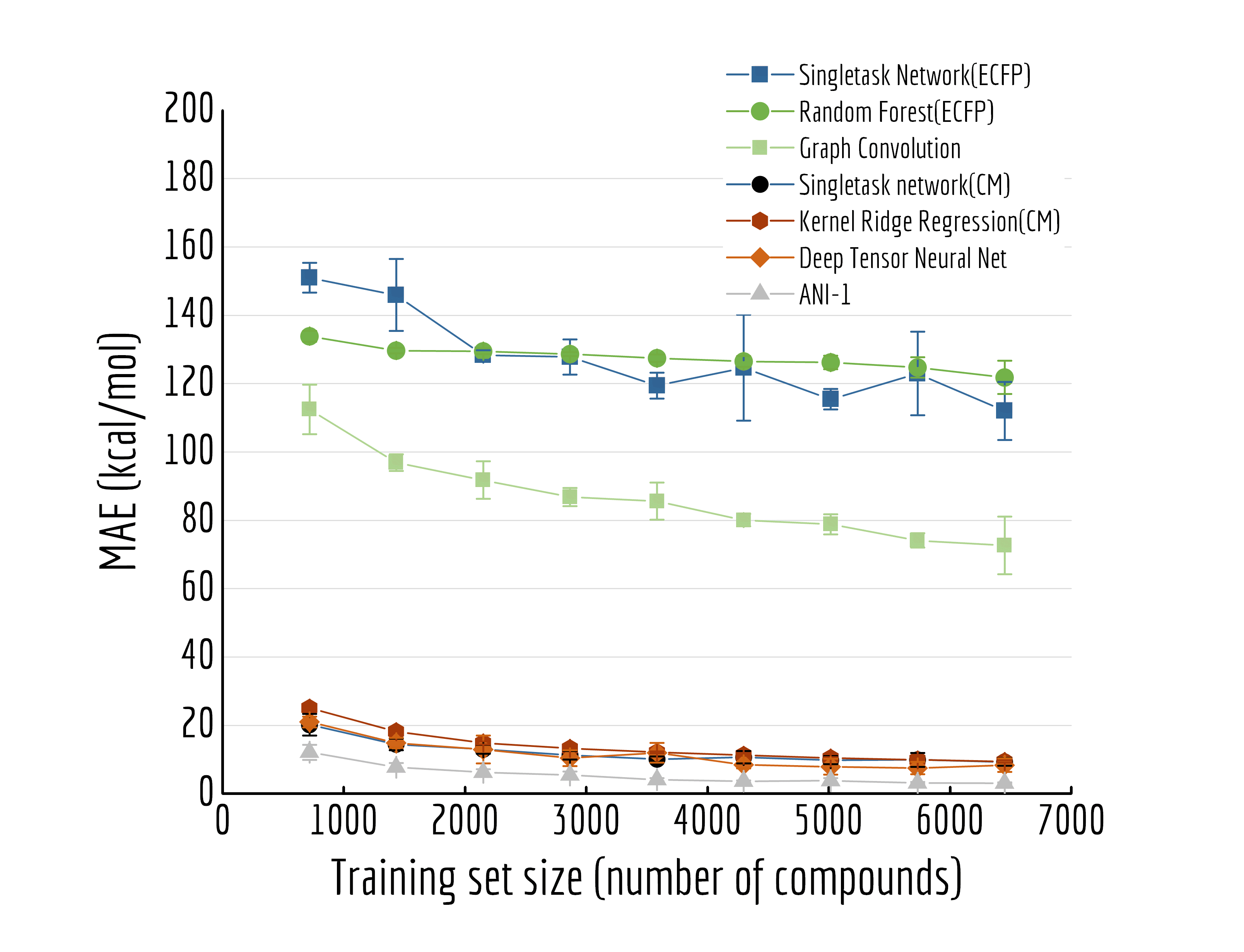}
  \caption{Out-of-sample performances with different training set sizes on QM7. Each datapoint is the average of 5 independent runs, with standard deviations shown as error bars.}
  \label{fig:QM7_variable}
\end{figure}

Another variable training size experiment is performed on QM7: predicting atomization energy. All mean absolute error performances are displayed in Figure~\ref{fig:QM7_variable}. Clearly incorporation of spatial position creates the huge gap between models, DTNN and multitask networks(CM) reach similar level of accuracy as reported in previous work on this dataset. (There is still a gap between the MoleculeNet implementation and best reported numbers from previous work\cite{GDB7_dataset_arxiv, schutt2016quantum}, which should be closed by training models longer, as indicated in Appendix, model validation part). ANI-1 reached the best performance on this task, illustrating overall lower mean absolute errors.

For QM series, proper choice of featurization appears critical. As mentioned previously, ECFP only consider graph substructures, while Coulomb Matrix and graph featurizations used by ANI-1, DTNN and MPNN are explicitly calculated on charges and physical distances, which are exactly the required inputs for solving Schr\"odinger's equation.

\section{Conclusion}

\begin{table}[h]
    \caption{Summary of performances(test subset): conventional methods versus graph-based methods. Graph-based models outperform conventional methods on 11/17 datasets.}
    \centering
    \footnotesize	
    \label{tab:conventional_versus_graph}
    \begin{threeparttable}[b]
    \begin{tabular*}{\textwidth}{@{\extracolsep{\fill}}lllll}
    \hline
    \multirow{2}{*}{\textbf{Category}} & \multirow{2}{*}{\textbf{Dataset}} & \multirow{2}{*}{\textbf{Metric}} & \textbf{Best performances - } & \textbf{Best performances - }\\
    & & & \textbf{conventional methods} & \textbf{graph-based methods} \\
    \hline
    \multirow{4}{*}{Quantum Mechanics} & QM7 & MAE & KRR(CM): 10.22 & \textbf{ANI-1: 2.86} \\\cline{2-5}
    & QM7b & MAE & KRR(CM): 1.05 & \textbf{DTNN: 1.77*} \\\cline{2-5}
    & QM8 & MAE & Multitask: 0.0150 & \textbf{MPNN: 0.0143} \\\cline{2-5}
    & QM9 & MAE & Multitask(CM): 4.35 & \textbf{DTNN: 2.35} \\\cline{2-5}
    \hline
    \multirow{3}{*}{Physical Chemistry} & ESOL & RMSE & XGBoost: 0.99 & \textbf{MPNN: 0.58} \\\cline{2-5}
    & FreeSolv & RMSE & XGBoost: 1.74 & \textbf{MPNN: 1.15} \\\cline{2-5}
    & Lipophilicity & RMSE & XGBoost: 0.799 & \textbf{GC: 0.655} \\
    \hline
    \multirow{5}{*}{Biophysics} & PCBA & AUC-PRC & Logreg: 0.129 & \textbf{GC: 0.136} \\\cline{2-5}
    & MUV & AUC-PRC & \textbf{Multitask: 0.184} & Weave: 0.109 \\\cline{2-5}
    & HIV & AUC-ROC & \textbf{KernelSVM: 0.792} & GC: 0.763 \\\cline{2-5}
    & BACE & AUC-ROC & \textbf{RF: 0.867} & Weave: 0.806 \\\cline{2-5}
    & PDBbind(full) & RMSE & \textbf{RF(grid): 1.25} & GC: 1.44 \\
    \hline
    \multirow{5}{*}{Physiology} & BBBP & AUC-ROC & \textbf{KernelSVM: 0.729} & GC: 0.690 \\\cline{2-5}
    & Tox21 & AUC-ROC & KernelSVM: 0.822 & \textbf{GC: 0.829} \\\cline{2-5}
    & ToxCast & AUC-ROC & Multitask: 0.702 & \textbf{Weave: 0.742} \\\cline{2-5}
    & SIDER & AUC-ROC & \textbf{RF: 0.684} & GC: 0.638 \\\cline{2-5}
    & ClinTox & AUC-ROC& Bypass: 0.827 & \textbf{Weave: 0.832} \\
    \hline
    \end{tabular*}    
    \begin{tablenotes}
        \item * As discussed in section 4.4, DTNN outperforms KRR(CM) on 14/16 tasks in QM7b while the mean-MAE is skewed due to different magnitudes of labels.
    \end{tablenotes}
    \end{threeparttable}
\end{table}

This work introduces MoleculeNet, a benchmark for molecular machine learning. We gathered data for a wide range of molecular properties: 17 dataset collections including over 800 different tasks on 700,000 compounds. Tasks are categorized into 4 levels as illustrated in Figure~\ref{fig:dataset_composition}: (i) quantum mechanical properties; (ii) physical chemistry properties; (iii) biophysical affinity and activity with bio-macromolecules; (iv) macroscopic physiological effects on human body. 

MoleculeNet contributes a data-loading framework, featurization methods, data splitting methods, and learning models to the open source DeepChem package (Figure~\ref{fig:benchmark_example}). By adding interchangeable featurizations, splits and learning models into the DeepChem framework, we can apply these primitives to the wide range of datasets in MoleculeNet. 

Broadly, our results show that graph-based models outperformed other methods by comfortable margins on most datasets(11/17, best performances comparison in Table~\ref{tab:conventional_versus_graph}), revealing a clear advantage of learnable featurizations. However, this effect has some caveats: Graph-based methods are not robust enough on complex tasks under data scarcity; on heavily imbalanced classification datasets, conventional methods such as kernel SVM outperform learnable featurizations with respect to recall of positives. Furthermore, for the PDBBind and quantum mechanics datasets, the use of appropriate featurizations which contain pertinent information is very significant. Comparing fully connected neural networks, random forests, and other comparatively simple algorithms, we claim that the PDBbind and QM7 results emphasize the necessity of using specialized features for different tasks. DTNN and MPNN which use distance information perform better on QM datasets than simple graph convolutions. While out of the scope of this paper, we note similarly that customized deep learning algorithms \cite{AtomNet} could in principle supplant the need for hand-derived, specialized features in such biophysical settings. On the FreeSolv dataset, comparison between conventional \textit{ab-initio} calculations and graph-based models for the prediction of solvation energies shows that data-driven methods can outperform physical algorithms with moderate amounts of data. These results suggest that data-driven physical chemistry will become increasingly important as methods mature. Results for biophysical and physiological datasets are currently weaker than for other datasets, suggesting that better featurizations or more data may be required for data-driven physiology to become broadly useful.


By providing a uniform platform for comparison and evaluation, we hope MoleculeNet will facilitate the development of new methods for both chemistry and machine learning. In future work, we hope to extend MoleculeNet to cover a broader range of molecular properties than considered here. For example, 3D protein structure prediction, or DNA topological modeling would benefit from the presence of strong benchmarks to encourage algorithmic development. We hope that the open-source design of MoleculeNet will encourage researchers to contribute implementations of other novel algorithms to the benchmark suite. In time, we hope to see MoleculeNet grow into a comprehensive resource for the molecular machine learning community.

\acknowledgement
We would like to thank the Stanford Computing Resources for providing us with access to the Sherlock and Xstream GPU nodes. Thanks to Steven Kearnes and Patrick Riley for early discussions about the MoleculeNet concept. Thanks to Aarthi Ramsundar for help with diagram construction.

Thanks to Zheng Xu for feedback on the MoleculeNet API. Thanks to Patrick Hop for contribution of the Lipophilicity dataset to MoleculeNet. Thanks to Anthony Gitter and Johnny Israeli for suggesting the addition of AuPRC for imbalanced datasets.

The Pande Group is broadly supported by grants from the NIH (R01 GM062868 and U19 AI109662) as well as gift funds and contributions from Folding@home donors.

We acknowledge the generous support of Dr. Anders G. Fr{\o}seth and Mr. Christian Sundt for our work on machine learning.

B.R. was supported by the Fannie and John Hertz Foundation.

\section{Appendix}

\subsection{Model Training and Hyperparameter Optimization}
All models were trained on Stanford's GPU clusters via DeepChem. No model was allowed to train for more than 10 hours(time profile in Table~\ref{tab:running_time}. Users can reproduce benchmarks locally by following directions from DeepChem.

Hyperparameters were determined using Gaussian Process Optimization via pyGPGO (https://github.com/hawk31/pyGPGO), with max number of iterations set to 20. Optimized hyperparameters for each model are listed, detailed hyperparameters can be found on Deepchem.

\subsubsection{Logistic Regression (Logreg)}
\begin{itemize}
    \item Learning rate
    \item L2 regularization
    \item Batch size
\end{itemize}

\subsubsection{Support Vector Classification (KernelSVM)}
\begin{itemize}
    \item Penalty parameter C
    \item Kernel coefficient gamma for radial basis function
\end{itemize}

\subsubsection{Kernel Ridge Regression (KRR)}
\begin{itemize}
    \item Penalty parameter
\end{itemize}

\subsubsection{Random Forest (RF)}
\begin{itemize}
    \item Number of trees in the forest: 500
\end{itemize}

\subsubsection{Gradient Boosting (XGBoost)}
\begin{itemize}
    \item Maximum tree depth
    \item Learning rate
    \item Number of boosted tree
\end{itemize}

\subsubsection{Multitask/Singletask Networks}
\begin{itemize}
    \item Layer size
    \item Weight - initial standard deviation
    \item Bias - initial constant
    \item Learning rate
    \item L2 regularization
    \item Batch size
\end{itemize}

\subsubsection{Bypass Networks}
\begin{itemize}
    \item Layer size(main layer and bypass layer)
    \item Weight - initial standard deviation(main layer and bypass layer)
    \item Bias - initial constant(main layer and bypass layer)
    \item Learning rate
    \item L2 regularization
    \item Batch size
\end{itemize}

\subsubsection{Influence Relevance Voting (IRV)}
\begin{itemize}
    \item K(number of nearest neighbors)
    \item Learning rate
    \item Batch size
\end{itemize}

\subsubsection{Graph Convolutional models (GC)}
\begin{itemize}
    \item Layer size of convolutional layers
    \item Layer size of fully-connected layer
    \item Learning rate
    \item Batch size
\end{itemize}

\subsubsection{Weave models}
\begin{itemize}
    \item Length of output features(layer size) of convolutional layers
    \item Learning rate
    \item Batch size
\end{itemize}

\subsubsection{Deep Tensor Neural Networks (DTNN)}
\begin{itemize}
    \item Length of atom embedding(features)
    \item Size of distance bin(from -1\AA~to 19\AA)
    \item Learning rate
    \item Batch size
\end{itemize}

\subsubsection{Directed Acyclic Graph models (DAG)}
\begin{itemize}
    \item Length of features in the convolutional layer
    \item Maximum number of propagation of a graph
    \item Learning rate
    \item Batch size
\end{itemize}

\subsubsection{Message Passing Neural Networks (MPNN)}
\begin{itemize}
    \item Number of message passing phases
    \item Number of steps(iterations) in readout phase
    \item Learning rate
    \item Batch size
\end{itemize}

\subsubsection{ANI-1}
\begin{itemize}
    \item Layer size
    \item Length of radial and angular symmetry functions
    \item Learning rate
    \item Batch size
\end{itemize}

All final performances were run three times with different fixed numerical seeds on the best-performing hyperparameters, and data splitting methods have been set to maintain determistic behavior. These settings control most randomness in learning process, but benchmark runs(on the same seed) may vary on the order of 1\% due to other sources of nondeterminism. Mean and standard deviations of all results are presented in the Performances section of Appendix.

We measured model running time of Tox21, MUV, QM8 and Lipophicility on a single node in Stanford's GPU clusters(CPU: Intel Xeon E5-2640 v3 @2.60 GHz, GPU: NVIDIA Tesla K80), results listed below:
\begin{table}[H]
    \small
    \centering
    \begin{threeparttable}
    \caption{Time Profile for Tox21, MUV, QM8 and Lipophilicity(second)}
    \begin{tabular}{ |c|c|c|c|c| } 
    \hline
    \textbf{Model} & \textbf{Tox21} & \textbf{MUV} & \textbf{QM8} & \textbf{Lipophilicity} \\
    \hline
    \hline
    Logreg & 93 & 522 & &\\
    \hline
    KernelSVM & 2574 & 2231 & & \\
    \hline
    KRR & & & 3390/5153* & 24\\
    \hline
    RF & 24273 & & & 186\\
    \hline
    XGBoost & 2082 & 2418 & & 410\\
    \hline
    Multitask/Singletask & 22 & 858 & 275/701* & 21\\
    \hline
    Bypass & 31 & 938 & &\\
    \hline
    IRV & 58 & 2674 & &\\
    \hline
    GC & 246 & 2320 & 512 & 131\\
    \hline
    Weave & 323 & 4593 & & 255\\
    \hline
    DAG & & & & 5142\\
    \hline
    DTNN & & & 940 &\\
    \hline
    MPNN & & & 3383 & 1626\\
    \hline
    \end{tabular}
    \label{tab:running_time}
    \begin{tablenotes}
        \item * ECFP/Coulomb Matrix
    \end{tablenotes}
    \end{threeparttable}
\end{table}

\subsection{Performances}
\begin{table}[H]
    \small
    \centering
    \caption{PCBA, MUV, HIV and BACE Performances: AUC-PRC for PCBA and MUV, AUC-ROC for HIV and BACE }
    \begin{tabular}{ |c|c|c|c|c| } 
    \hline
    \textbf{Model} & \textbf{Model} & \textbf{Training} & \textbf{Validation} & \textbf{Test} \\
    \hline
    \hline    
    \multirow{4}{*}{PCBA}
    & Logreg & $0.166\pm0.001$ & $0.130\pm0.004$ & $0.129\pm0.003$ \\\cline{2-5}
    & Multitask & $0.100\pm0.003$ & $0.097\pm0.000$ & $0.100\pm0.006$ \\\cline{2-5}
    & Bypass & $0.121\pm0.001$ & $0.111\pm0.003$ & $0.112\pm0.002$ \\\cline{2-5}
    & GC & $0.151\pm0.001$ & $\mathbf{0.136\pm0.003}$ & $\mathbf{0.136\pm0.004}$ \\\cline{2-5}
    \hline
    \hline    
    \multirow{8}{*}{MUV}
    & Logreg & $0.238\pm0.010$ & $0.036\pm0.009$ & $0.070\pm0.009$ \\\cline{2-5}
    & KernelSVM & $0.922\pm0.034$ & $0.113\pm0.039$ & $0.137\pm0.033$ \\\cline{2-5}
    & XGBoost & $0.159\pm0.018$ & $0.066\pm0.053$ & $0.086\pm0.033$ \\\cline{2-5}
    & IRV & $0.043\pm0.006$ & $0.069\pm0.008$ & $0.087\pm0.025$ \\\cline{2-5}
    & Multitask & $0.385\pm0.014$ & $\mathbf{0.202\pm0.032}$ & $\mathbf{0.184\pm0.020}$ \\\cline{2-5}
    & Bypass & $0.317\pm0.027$ & $0.166\pm0.043$ & $0.148\pm0.069$ \\\cline{2-5}
    & GC & $0.040\pm0.013$ & $0.049\pm0.023$ & $0.046\pm0.031$ \\\cline{2-5}
    & Weave & $0.060\pm0.030$ & $0.127\pm0.028$ & $0.109\pm0.028$ \\\cline{2-5}
    \hline
    \hline    
    \multirow{8}{*}{HIV}
    & Logreg & $0.834\pm0.004$ & $0.788\pm0.016$ & $0.702\pm0.018$ \\\cline{2-5}
    & KernelSVM & $0.999\pm0.000$ & $0.837\pm0.000$ & $\mathbf{0.792\pm0.000}$ \\\cline{2-5}
    & XGBoost & $0.942\pm0.000$ & $\mathbf{0.841\pm0.000}$ & $0.756\pm0.000$ \\\cline{2-5}
    & IRV & $0.849\pm0.000$ & $0.818\pm0.000$ & $0.737\pm0.000$ \\\cline{2-5}
    & Multitask & $0.753\pm0.012$ & $0.711\pm0.027$ & $0.698\pm0.037$ \\\cline{2-5}
    & Bypass & $0.736\pm0.017$ & $0.719\pm0.012$ & $0.693\pm0.026$ \\\cline{2-5}
    & GC & $0.903\pm0.004$ & $0.792\pm0.014$ & $0.763\pm0.016$ \\\cline{2-5}
    & Weave & $0.725\pm0.004$ & $0.742\pm0.040$ & $0.703\pm0.039$ \\\cline{2-5}
    \hline
    \hline    
    \multirow{9}{*}{BACE}
    & Logreg & $0.960\pm0.001$ & $0.719\pm0.003$ & $0.781\pm0.010$ \\\cline{2-5}
    & KernelSVM & $0.986\pm0.000$ & $0.739\pm0.000$ & $0.862\pm0.000$ \\\cline{2-5}
    & XGBoost & $0.933\pm0.000$ & $\mathbf{0.756\pm0.000}$ & $0.850\pm0.000$ \\\cline{2-5}
    & RF & $0.999\pm0.000$ & $0.728\pm0.004$ & $\mathbf{0.867\pm0.008}$ \\\cline{2-5}
    & IRV & $0.887\pm0.000$ & $0.715\pm0.001$ & $0.838\pm0.000$ \\\cline{2-5}
    & Multitask & $0.863\pm0.034$ & $0.696\pm0.037$ & $0.824\pm0.006$ \\\cline{2-5}
    & Bypass & $0.931\pm0.001$ & $0.745\pm0.017$ & $0.829\pm0.006$ \\\cline{2-5}
    & GC & $0.852\pm0.046$ & $0.627\pm0.015$ & $0.783\pm0.014$ \\\cline{2-5}
    & Weave & $0.862\pm0.009$ & $0.638\pm0.014$ & $0.806\pm0.002$ \\\cline{2-5}
    \hline
    \end{tabular}
    \label{tab:PCBA_MUV_HIV_BACE}
\end{table}
    
\begin{table}[H]
    \small
    \centering
    \caption{BBBP, Tox21, ToxCast, SIDER, ClinTox Performances (AUC-ROC)}
    \begin{tabular}{ |c|c|c|c|c| } 
    \hline
    \textbf{Model} & \textbf{Model} & \textbf{Training} & \textbf{Validation} & \textbf{Test} \\
    \hline
    \hline    
    \multirow{9}{*}{BBBP}
    & Logreg & $0.986\pm0.001$ & $0.958\pm0.003$ & $0.699\pm0.002$ \\\cline{2-5}
    & KernelSVM & $0.995\pm0.000$ & $0.964\pm0.000$ & $\mathbf{0.729\pm0.000}$ \\\cline{2-5}
    & XGBoost & $0.987\pm0.000$ & $0.956\pm0.000$ & $0.696\pm0.000$ \\\cline{2-5}
    & RF & $1.000\pm0.000$ & $0.956\pm0.002$ & $0.714\pm0.000$ \\\cline{2-5}
    & IRV & $0.915\pm0.000$ & $\mathbf{0.964\pm0.000}$ & $0.700\pm0.000$ \\\cline{2-5}
    & Multitask & $0.908\pm0.019$ & $0.955\pm0.002$ & $0.688\pm0.005$ \\\cline{2-5}
    & Bypass & $0.950\pm0.005$ & $0.960\pm0.003$ & $0.702\pm0.006$ \\\cline{2-5}
    & GC & $0.956\pm0.004$ & $0.943\pm0.002$ & $0.690\pm0.009$ \\\cline{2-5}
    & Weave & $0.873\pm0.010$ & $0.951\pm0.005$ & $0.671\pm0.014$ \\\cline{2-5}
    \hline
    \hline    
    \multirow{9}{*}{Tox21}
    & Logreg & $0.910\pm0.002$ & $0.772\pm0.011$ & $0.794\pm0.015$ \\\cline{2-5}
    & KernelSVM & $0.998\pm0.000$ & $0.818\pm0.010$ & $0.822\pm0.006$ \\\cline{2-5}
    & XGBoost & $0.899\pm0.011$ & $0.775\pm0.018$ & $0.794\pm0.014$ \\\cline{2-5}
    & RF & $0.999\pm0.000$ & $0.763\pm0.002$ & $0.769\pm0.015$ \\\cline{2-5}
    & IRV & $0.805\pm0.003$ & $0.807\pm0.006$ & $0.799\pm0.006$ \\\cline{2-5}
    & Multitask & $0.884\pm0.001$ & $0.795\pm0.017$ & $0.803\pm0.012$ \\\cline{2-5}
    & Bypass & $0.938\pm0.001$ & $0.800\pm0.008$ & $0.810\pm0.013$ \\\cline{2-5}
    & GC & $0.905\pm0.004$ & $0.825\pm0.013$ & $\mathbf{0.829\pm0.006}$ \\\cline{2-5}
    & Weave & $0.875\pm0.004$ & $\mathbf{0.828\pm0.008}$ & $0.820\pm0.010$ \\\cline{2-5}
    \hline
    \hline    
    \multirow{8}{*}{ToxCast}
    & Logreg & $0.828\pm0.016$ & $0.611\pm0.024$ & $0.605\pm0.003$ \\\cline{2-5}
    & KernelSVM & $0.905\pm0.012$ & $0.674\pm0.013$ & $0.669\pm0.014$ \\\cline{2-5}
    & XGBoost & $0.764\pm0.004$ & $0.641\pm0.009$ & $0.640\pm0.005$ \\\cline{2-5}
    & IRV & $0.663\pm0.004$ & $0.660\pm0.009$ & $0.663\pm0.015$ \\\cline{2-5}
    & Multitask & $0.887\pm0.002$ & $0.705\pm0.017$ & $0.702\pm0.013$ \\\cline{2-5}
    & Bypass & $0.793\pm0.002$ & $0.684\pm0.016$ & $0.676\pm0.005$ \\\cline{2-5}
    & GC & $0.815\pm0.003$ & $0.709\pm0.013$ & $0.716\pm0.014$ \\\cline{2-5}
    & Weave & $0.830\pm0.006$ & $\mathbf{0.750\pm0.007}$ & $\mathbf{0.742\pm0.003}$ \\\cline{2-5}
    \hline
    \hline    
    \multirow{9}{*}{SIDER}
    & Logreg & $0.918\pm0.001$ & $0.635\pm0.018$ & $0.643\pm0.011$ \\\cline{2-5}
    & KernelSVM & $0.984\pm0.021$ & $0.655\pm0.030$ & $0.682\pm0.013$ \\\cline{2-5}
    & XGBoost & $0.854\pm0.016$ & $0.645\pm0.038$ & $0.656\pm0.027$ \\\cline{2-5}
    & RF & $1.000\pm0.000$ & $0.650\pm0.013$ & $\mathbf{0.684\pm0.009}$ \\\cline{2-5}
    & IRV & $0.628\pm0.004$ & $\mathbf{0.657\pm0.028}$ & $0.640\pm0.020$ \\\cline{2-5}
    & Multitask & $0.790\pm0.007$ & $0.632\pm0.040$ & $0.666\pm0.026$ \\\cline{2-5}
    & Bypass & $0.852\pm0.001$ & $0.644\pm0.035$ & $0.673\pm0.025$ \\\cline{2-5}
    & GC & $0.735\pm0.013$ & $0.609\pm0.021$ & $0.638\pm0.012$ \\\cline{2-5}
    & Weave & $0.647\pm0.015$ & $0.591\pm0.031$ & $0.581\pm0.027$ \\\cline{2-5}
    \hline
    \hline    
    \multirow{9}{*}{ClinTox}
    & Logreg & $0.990\pm0.001$ & $0.732\pm0.065$ & $0.722\pm0.039$ \\\cline{2-5}
    & KernelSVM & $0.994\pm0.002$ & $0.614\pm0.195$ & $0.669\pm0.092$ \\\cline{2-5}
    & XGBoost & $0.926\pm0.008$ & $0.729\pm0.140$ & $0.799\pm0.050$ \\\cline{2-5}
    & RF & $0.996\pm0.001$ & $0.688\pm0.107$ & $0.713\pm0.056$ \\\cline{2-5}
    & IRV & $0.804\pm0.004$ & $0.748\pm0.075$ & $0.770\pm0.072$ \\\cline{2-5}
    & Multitask & $0.917\pm0.002$ & $0.711\pm0.186$ & $0.778\pm0.055$ \\\cline{2-5}
    & Bypass & $0.943\pm0.004$ & $0.734\pm0.111$ & $0.827\pm0.051$ \\\cline{2-5}
    & GC & $0.962\pm0.005$ & $\mathbf{0.920\pm0.035}$ & $0.807\pm0.047$ \\\cline{2-5}
    & Weave & $0.948\pm0.013$ & $0.875\pm0.066$ & $\mathbf{0.832\pm0.037}$ \\\cline{2-5}
    \hline
    \end{tabular}
    \label{tab:BBBP_Tox21_ToxCast_SIDER}
\end{table}
    
\begin{table}[H]
    \small
    \centering
    \caption{PDBbind Performances (Root-Mean-Square Error)}
    \begin{tabular}{ |c|c|c|c|c| } 
    \hline
    \textbf{Model} & \textbf{Model} & \textbf{Training} & \textbf{Validation} & \textbf{Test} \\
    \hline
    \hline    
    \multirow{5}{*}{PDBbind - core}
    & RF & $0.82\pm0.00$ & $2.02\pm0.02$ & $2.03\pm0.01$ \\\cline{2-5}
    & RF(grid) & $0.73\pm0.01$ & $1.98\pm0.01$ & $2.27\pm0.01$ \\\cline{2-5}
    & Multitask & $1.62\pm0.03$ & $\mathbf{1.86\pm0.01}$ & $2.21\pm0.02$ \\\cline{2-5}
    & Multitask(grid) & $1.51\pm0.05$ & $1.92\pm0.02$ & $2.20\pm0.03$ \\\cline{2-5}
    & GC & $1.42\pm0.04$ & $2.10\pm0.05$ & $\mathbf{1.92\pm0.07}$ \\\cline{2-5}
    \hline
    \hline    
    \multirow{5}{*}{PDBbind - refined}
    & RF & $0.66\pm0.00$ & $1.48\pm0.00$ & $1.62\pm0.00$ \\\cline{2-5}
    & RF(grid) & $0.51\pm0.00$ & $\mathbf{1.37\pm0.00}$ & $\mathbf{1.38\pm0.00}$ \\\cline{2-5}
    & Multitask & $1.09\pm0.01$ & $1.53\pm0.03$ & $1.66\pm0.05$ \\\cline{2-5}
    & Multitask(grid) & $0.55\pm0.02$ & $1.41\pm0.02$ & $1.46\pm0.05$ \\\cline{2-5}
    & GC & $1.20\pm0.01$ & $1.55\pm0.05$ & $1.65\pm0.03$ \\\cline{2-5}
    \hline
    \hline    
    \multirow{5}{*}{PDBbind - full}
    & RF & $0.66\pm0.00$ & $1.40\pm0.00$ & $1.31\pm0.00$ \\\cline{2-5}
    & RF(grid) & $0.51\pm0.00$ & $\mathbf{1.35\pm0.00}$ & $\mathbf{1.25\pm0.00}$ \\\cline{2-5}
    & Multitask & $1.52\pm0.17$ & $1.42\pm0.05$ & $1.45\pm0.14$ \\\cline{2-5}
    & Multitask(grid) & $0.39\pm0.01$ & $1.40\pm0.03$ & $1.28\pm0.02$ \\\cline{2-5}
    & GC & $1.65\pm0.10$ & $1.57\pm0.20$ & $1.44\pm0.12$ \\\cline{2-5}
    \hline
    \end{tabular}
    \label{tab:PDBbind}
\end{table}

\begin{table}[H]
    \small
    \centering
    \caption{ESOL, FreeSolv, Lipophilicity Performances (Root-Mean-Square Error)}
    \begin{tabular}{ |c|c|c|c|c| } 
    \hline
    \textbf{Model} & \textbf{Model} & \textbf{Training} & \textbf{Validation} & \textbf{Test} \\
    \hline
    \hline    
    \multirow{8}{*}{ESOL}
    & RF & $0.51\pm0.01$ & $1.16\pm0.15$ & $1.07\pm0.19$ \\\cline{2-5}
    & Multitask & $0.59\pm0.04$ & $1.17\pm0.13$ & $1.12\pm0.15$ \\\cline{2-5}
    & XGBoost & $0.51\pm0.08$ & $1.05\pm0.10$ & $0.99\pm0.14$ \\\cline{2-5}
    & KRR & $0.38\pm0.01$ & $1.65\pm0.19$ & $1.53\pm0.06$ \\\cline{2-5}
    & GC & $0.43\pm0.20$ & $1.05\pm0.15$ & $0.97\pm0.01$ \\\cline{2-5}
    & DAG & $0.32\pm0.03$ & $0.74\pm0.04$ & $0.82\pm0.08$ \\\cline{2-5}
    & Weave & $0.34\pm0.04$ & $0.57\pm0.04$ & $0.61\pm0.07$ \\\cline{2-5}
    & MPNN & $0.25\pm0.06$ & $\mathbf{0.55\pm0.02}$ & $\mathbf{0.58\pm0.03}$ \\\cline{2-5}
    \hline
    \hline    
    \multirow{8}{*}{FreeSolv}
    & RF & $0.80\pm0.03$ & $2.12\pm0.68$ & $2.03\pm0.22$ \\\cline{2-5}
    & Multitask & $1.07\pm0.06$ & $1.95\pm0.41$ & $1.87\pm0.07$ \\\cline{2-5}
    & XGBoost & $0.85\pm0.12$ & $1.76\pm0.21$ & $1.74\pm0.15$ \\\cline{2-5}
    & KRR & $0.31\pm0.03$ & $2.10\pm0.12$ & $2.11\pm0.07$ \\\cline{2-5}
    & GC & $0.31\pm0.09$ & $1.35\pm0.15$ & $1.40\pm0.16$ \\\cline{2-5}
    & DAG & $0.49\pm0.46$ & $1.48\pm0.15$ & $1.63\pm0.18$ \\\cline{2-5}
    & Weave & $0.32\pm0.04$ & $\mathbf{1.19\pm0.08}$ & $1.22\pm0.28$ \\\cline{2-5}
    & MPNN & $0.31\pm0.05$ & $1.20\pm0.02$ & $\mathbf{1.15\pm0.12}$ \\\cline{2-5}
    \hline
    \hline    
    \multirow{8}{*}{Lipophilicity}
    & RF & $0.318\pm0.006$ & $0.835\pm0.036$ & $0.876\pm0.040$ \\\cline{2-5}
    & Multitask & $0.385\pm0.065$ & $0.852\pm0.048$ & $0.859\pm0.013$ \\\cline{2-5}
    & XGBoost & $0.135\pm0.012$ & $0.783\pm0.021$ & $0.799\pm0.054$ \\\cline{2-5}
    & KRR & $0.180\pm0.002$ & $0.889\pm0.009$ & $0.899\pm0.043$ \\\cline{2-5}
    & GC & $0.471\pm0.001$ & $\mathbf{0.678\pm0.040}$ & $\mathbf{0.655\pm0.036}$ \\\cline{2-5}
    & DAG & $0.173\pm0.026$ & $0.857\pm0.050$ & $0.835\pm0.039$ \\\cline{2-5}
    & Weave & $0.549\pm0.051$ & $0.734\pm0.011$ & $0.715\pm0.035$ \\\cline{2-5}
    & MPNN & $0.363\pm0.043$ & $0.757\pm0.030$ & $0.719\pm0.031$ \\\cline{2-5}
    \hline
    \end{tabular}
    \label{tab:ESOL_FreeSolv_Lipo}
\end{table}

\begin{table}[H]
    \small
    \centering
    \caption{QM7, QM7b, QM8 and QM9 Performances (Mean Absolute Error)}
    \begin{tabular}{ |c|c|c|c|c| } 
    \hline
    \textbf{Model} & \textbf{Model} & \textbf{Training} & \textbf{Validation} & \textbf{Test} \\
    \hline
    \hline    
    \multirow{8}{*}{QM7}
    & RF & $47.1\pm0.1$ & $124.0\pm4.6$ & $122.7\pm4.2$ \\\cline{2-5}
    & Multitask & $101.8\pm13.7$ & $121.7\pm7.5$ & $123.7\pm15.6$ \\\cline{2-5}
    & KRR & $65.5\pm0.3$ & $108.3\pm5.4$ & $110.3\pm4.7$ \\\cline{2-5}
    & GC & $67.8\pm4.0$ & $77.9\pm10.0$ & $77.9\pm2.1$ \\\cline{2-5}
    & Multitask(CM) & $10.4\pm1.8$ & $11.0\pm1.7$ & $10.8\pm1.3$ \\\cline{2-5}
    & KRR(CM) & $0.1\pm0.0$ & $9.9\pm0.1$ & $10.2\pm0.3$ \\\cline{2-5}
    & DTNN & $8.2\pm3.9$ & $8.9\pm3.7$ & $8.8\pm3.5$ \\\cline{2-5}
    & ANI-1 & $2.42\pm0.32$ & $\mathbf{2.99\pm0.22}$ & $\mathbf{2.86\pm0.25}$ \\\cline{2-5}
    \hline
    \hline
    \multirow{3}{*}{QM7b}
    & Multitask(CM) & $2.95\pm0.70$ & $2.90\pm0.82$ & $2.89\pm0.65$ \\\cline{2-5}
    & KRR(CM) & $0.01\pm0.00$ & $\mathbf{1.08\pm0.08}$ & $\mathbf{1.05\pm0.06}$ \\\cline{2-5}
    & DTNN & $1.68\pm0.18$ & $1.79\pm0.14$ & $1.77\pm0.17$ \\\cline{2-5}
    \hline
    \hline    
    \multirow{7}{*}{QM8}
    & Multitask & $0.0081\pm0.0002$ & $0.0155\pm0.0005$ & $0.0150\pm0.0005$ \\\cline{2-5}
    & KRR & $0.0152\pm0.0001$ & $0.0197\pm0.0004$ & $0.0195\pm0.0003$ \\\cline{2-5}
    & GC & $0.0123\pm0.0009$ & $0.0150\pm0.0006$ & $0.0148\pm0.0006$ \\\cline{2-5}
    & Multitask(CM) & $0.0163\pm0.0010$ & $0.0181\pm0.0012$ & $0.0179\pm0.0013$ \\\cline{2-5}
    & KRR(CM) & $0.0002\pm0.0000$ & $0.0242\pm0.0003$ & $0.0238\pm0.0004$ \\\cline{2-5}
    & DTNN & $0.0140\pm0.0009$ & $0.0170\pm0.0007$ & $0.0169\pm0.0009$ \\\cline{2-5}
    & MPNN & $0.0128\pm0.0010$ & $\mathbf{0.0146\pm0.0010}$ & $\mathbf{0.0143\pm0.0011}$ \\\cline{2-5}
    \hline
    \hline    
    \multirow{5}{*}{QM9}
    & Multitask & $15.3\pm0.2$ & $15.9\pm0.2$ & $16.0\pm0.2$ \\\cline{2-5}
    & GC & $4.6\pm0.5$ & $4.7\pm0.5$ & $4.7\pm0.5$ \\\cline{2-5}
    & Multitask(CM) & $4.3\pm1.0$ & $4.3\pm1.1$ & $4.4\pm1.0$ \\\cline{2-5}
    & DTNN & $2.3\pm1.1$ & $\mathbf{2.4\pm1.1}$ & $\mathbf{2.4\pm1.1}$ \\\cline{2-5}
    & MPNN & $3.2\pm1.5$ & $3.2\pm1.5$ & $3.2\pm1.5$ \\\cline{2-5}
    \hline
    \end{tabular}
    \label{tab:QM7_QM7b_QM8_QM9}
\end{table}

\begin{table}[H]
    \small
    \centering
    \caption{QM7b Test Set Performances of All Tasks(Mean Absolute Error)}
    \begin{tabular}{ |c|c|c|c| } 
    \hline
    \textbf{Task} & Multitask(CM) & KRR(CM) & DTNN \\
    \hline
    \hline    
    Atomization energy - PBE0 & $36.0$ & $\mathbf{9.3}$ & $21.5$\\
    \hline    
    Excitation energy of maximal optimal absorption - ZINDO & $1.31$ & $1.83$ & $\mathbf{1.26}$\\
    \hline    
    Highest absorption - ZINDO & $0.086$ & $0.098$ & $\mathbf{0.074}$\\
    \hline
    HOMO - ZINDO & $0.293$ & $0.369$ & $\mathbf{0.192}$\\
    \hline
    LUMO - ZINDO & $0.255$ & $0.361$ & $\mathbf{0.159}$\\
    \hline
    1st excitation energy - ZINDO & $0.368$ & $0.479$ & $\mathbf{0.296}$\\
    \hline
    Ionization potential - ZINDO & $0.305$ & $0.408$ & $\mathbf{0.214}$\\
    \hline
    Electron Affinity - ZINDO & $0.271$ & $0.404$ & $\mathbf{0.174}$\\
    \hline
    HOMO - KS & $0.247$ & $0.272$ & $\mathbf{0.155}$\\
    \hline
    LUMO - KS & $0.187$ & $0.239$ & $\mathbf{0.129}$\\
    \hline
    HOMO - GW & $0.270$ & $0.294$ & $\mathbf{0.166}$\\
    \hline
    LUMO - GW & $0.172$ & $0.236$ & $\mathbf{0.139}$\\
    \hline
    Polarizability - PBE0 & $0.335$ & $0.225$ & $\mathbf{0.173}$\\
    \hline
    Polarizability - SCS & $0.317$ & $\mathbf{0.116}$ & $0.149$\\
    \hline
    \end{tabular}
    \label{tab:QM7b}
\end{table}

\begin{table}[H]
    \small
    \centering
    \caption{QM8 Test Set Performances of All Tasks(Mean Absolute Error)}
    \begin{tabular}{ |c|c|c|c|c|c|c|c| } 
    \hline
    \textbf{Task} & Multitask & GC & KRR & Multitask(CM) & KRR(CM) & DTNN & MPNN\\
    \hline
    \hline    
    E1 - CC2 & $0.0088$ & $\mathbf{0.0074}$ & $0.0115$ & $0.0125$ & $0.0137$ & $0.0092$ & $0.0084$\\
    \hline    
    E2 - CC2 & $0.0098$ & $\mathbf{0.0085}$ & $0.0116$ & $0.0114$ & $0.0124$ & $0.0092$ & $0.0091$\\
    \hline    
    f1 - CC2 & $\mathbf{0.0145}$ & $0.0175$ & $0.0202$ & $0.0186$ & $0.0272$ & $0.0182$ & $0.0151$\\
    \hline
    f2 - CC2 & $0.0320$ & $0.0328$ & $0.0387$ & $0.0358$ & $0.0460$ & $0.0377$ & $\mathbf{0.0314}$\\
    \hline    
    E1 - PBE0 & $0.0089$ & $\mathbf{0.0076}$ & $0.0118$ & $0.0126$ & $0.0140$ & $0.0090$ & $0.0083$\\
    \hline    
    E2 - PBE0 & $0.0096$ & $\mathbf{0.0083}$ & $0.0117$ & $0.0114$ & $0.0122$ & $0.0086$ & $0.0086$\\
    \hline    
    f1 - PBE0 & $\mathbf{0.0121}$ & $0.0125$ & $0.0189$ & $0.0152$ & $0.0258$ & $0.0155$ & $0.0123$\\
    \hline
    f2 - PBE0 & $0.0252$ & $0.0246$ & $0.0319$ & $0.0267$ & $0.0376$ & $0.0281$ & $\mathbf{0.0236}$\\
    \hline    
    E1 - CAM & $0.0083$ & $\mathbf{0.0070}$ & $0.0111$ & $0.0119$ & $0.0132$ & $0.0086$ & $0.0079$\\
    \hline    
    E2 - CAM & $0.0090$ & $\mathbf{0.0076}$ & $0.0109$ & $0.0106$ & $0.0115$ & $0.0082$ & $0.0082$\\
    \hline    
    f1 - CAM & $0.0140$ & $0.0153$ & $0.0208$ & $0.0177$ & $0.0304$ & $0.0180$ & $\mathbf{0.0134}$\\
    \hline
    f2 - CAM & $0.0274$ & $0.0285$ & $0.0345$ & $0.0303$ & $0.0417$ & $0.0322$ & $\mathbf{0.0258}$\\
    \hline
    \end{tabular}
    \label{tab:QM8}
\end{table}

\begin{table}[H]
    \small
    \centering
    \caption{QM9 Test Set Performances of All Tasks(Mean Absolute Error)}
    \begin{tabular}{ |c|c|c|c|c|c| } 
    \hline
    \textbf{Task} & Multitask & Multitask(CM) & GC  & DTNN & MPNN\\
    \hline
    \hline    
    mu  & $0.602$ & $0.519$ & $0.583$ & $\mathbf{0.244}$ & $0.358$\\
    \hline    
    alpha  & $3.10$ & $\mathbf{0.85}$ & $1.37$ & $0.95$ & $0.89$\\
    \hline    
    HOMO  & $0.00660$ & $0.00506$ & $0.00716$ & $\mathbf{0.00388}$ & $0.00541$\\
    \hline
    LUMO  & $0.00854$ & $0.00645$ & $0.00921$ & $\mathbf{0.00513}$ & $0.00623$\\
    \hline    
    gap  & $0.0100$ & $0.0086$ & $0.0112$ & $\mathbf{0.0066}$ & $0.0082$\\
    \hline    
    R2  & $125.7$ & $46.0$ & $35.9$ & $\mathbf{17.0}$ & $28.5$\\
    \hline    
    ZPVE  & $0.01109$ & $0.00207$ & $0.00299$ & $\mathbf{0.00172}$ & $0.00216$\\
    \hline
    U0  & $15.10$ & $2.27$ & $3.41$ & $2.43$ & $\mathbf{2.05}$\\
    \hline    
    U  & $15.10$ & $2.27$ & $3.41$ & $2.43$ & $\mathbf{2.00}$\\
    \hline    
    H  & $15.10$ & $2.27$ & $3.41$ & $2.43$ & $\mathbf{2.02}$\\
    \hline    
    G  & $15.10$ & $2.27$ & $3.41$ & $2.43$ & $\mathbf{2.02}$\\
    \hline
    Cv  & $1.77$ & $0.39$ & $0.65$ & $\mathbf{0.27}$ & $0.42$\\
    \hline
    \end{tabular}
    \label{tab:QM9}
\end{table}

\subsection{Grid Featurizer}

In our implementation, we generate a vector with length 2052 for each pair of ligand and protein. Detailed process listed below:

First, binding pocket atoms of the protein are extracted using a distance cutoff of 4.5 \AA. In this process, atom in the protein will be extracted only if it locates within this distance from any atom in the ligand molecule. 

Intra-ligand and intra-protein fingerprints are generated (using the ordinary circular fingerprint with radius of 2) respectively on the atoms from the ligand and atoms in the binding pocket of the protein, and then hashed together to form a vector of length 512.

Then we form three different sets of contacting atom pairs between ligand and protein, whose intra-pair distance falls within bins: $0\sim2$ \AA, $2\sim3$ \AA and $3\sim4.5$ \AA. Each set of pairs is hashed into a fixed length fingerprint with length 512.

Finally, salt bridges are counted, hydrogen bonds are counted in three different distance bins, forming the last four digits. In total the fingerprints have length of 2052.

\subsection{ClinTox}

The ClinTox dataset addresses clinical drug toxicity by providing a qualitative comparison of drugs approved by the FDA and those that have failed clinical trials for toxicity reasons. We compiled the FDA-approved drug names from annotations in the SWEETLEAD database. We compiled the names of drugs that failed clinical trials for toxicity reasons from the Aggregate Analysis of ClinicalTrials.gov (AACT) database. To identify these drug names, we relied on annotations from the clinical study table titled "clinical\_study\_noclob.txt" in the AACT database. From this table, we selected clinical trials where the overall status was "terminated," "suspended," or "withdrawn," and the explanation for the status included the terms "adverse," "toxic," or "death."

\subsection{Dataset and model access}

Table~\ref{tab:deepchem_command} listed DeepChem commands to load datasets and models in MoleculeNet. For more detailed instructions please refer to the docs and examples. Tutorial for building customized datasets can be found at \url{https://github.com/deepchem/deepchem/blob/master/examples/notebooks/dataset_preparation.ipynb}

\begin{table}[H]
    \centering
    \small
    \caption{DeepChem commands to load MoleculeNet datasets and models}
    \begin{tabular}{ |c|l| } 
    \hline
    \textbf{Dataset} & \textbf{Command}\\
    \hline
    QM7 & {\fontfamily{pcr}\selectfont deepchem.molnet.load\_qm7\_from\_mat}\\
    \hline
    QM7b & {\fontfamily{pcr}\selectfont deepchem.molnet.load\_qm7b\_from\_mat}\\
    \hline
    QM8 & {\fontfamily{pcr}\selectfont deepchem.molnet.load\_qm8}\\
    \hline
    QM9 & {\fontfamily{pcr}\selectfont deepchem.molnet.load\_qm9}\\
    \hline
    ESOL & {\fontfamily{pcr}\selectfont deepchem.molnet.load\_delaney}\\
    \hline
    FreeSolv & {\fontfamily{pcr}\selectfont deepchem.molnet.load\_sampl}\\
    \hline
    Lipophilicity & {\fontfamily{pcr}\selectfont deepchem.molnet.load\_lipo}\\
    \hline
    PCBA & {\fontfamily{pcr}\selectfont deepchem.molnet.load\_pcba}\\
    \hline
    MUV & {\fontfamily{pcr}\selectfont deepchem.molnet.load\_muv}\\
    \hline
    HIV & {\fontfamily{pcr}\selectfont deepchem.molnet.load\_hiv}\\
    \hline
    BACE & {\fontfamily{pcr}\selectfont deepchem.molnet.load\_bace\_classification}\\
    \hline
    PDBbind & {\fontfamily{pcr}\selectfont deepchem.molnet.load\_pdbbind\_grid}\\
    \hline
    BBBP & {\fontfamily{pcr}\selectfont deepchem.molnet.load\_bbbp}\\
    \hline
    Tox21 & {\fontfamily{pcr}\selectfont deepchem.molnet.load\_tox21}\\
    \hline
    ToxCast & {\fontfamily{pcr}\selectfont deepchem.molnet.load\_toxcast}\\
    \hline
    SIDER & {\fontfamily{pcr}\selectfont deepchem.molnet.load\_sider}\\
    \hline
    ClinTox & {\fontfamily{pcr}\selectfont deepchem.molnet.load\_clintox}\\
    \hline
    \hline
    \textbf{Model} & \textbf{Command} \\
    \hline
    Logreg$^a$ & {\fontfamily{pcr}\selectfont sklearn.linear\_model.LogisticRegression} \\
    \hline
    KernelSVM$^a$ & {\fontfamily{pcr}\selectfont sklearn.svm.SVC} \\
    \hline
    KRR$^a$ & {\fontfamily{pcr}\selectfont sklearn.kernel\_ridge.KernelRidge} \\
    \hline
    \multirow{2}{*}{RF$^a$} & {\fontfamily{pcr}\selectfont sklearn.ensemble.RandomForestClassifier} \\
    & {\fontfamily{pcr}\selectfont sklearn.ensemble.RandomForestRegressor} \\
    \hline
    XGBoost$^b$ & {\fontfamily{pcr}\selectfont deepchem.models.xgboost\_models.XGBoostModel} \\ 
    \hline
    \multirow{2}{*}{Multitask/Singletask} & {\fontfamily{pcr}\selectfont deepchem.models.MultitaskClassifier} \\
    & {\fontfamily{pcr}\selectfont deepchem.models.MultitaskRegressor} \\
    \hline
    Bypass & {\fontfamily{pcr}\selectfont deepchem.models.RobustMultitaskClassifier} \\
    \hline
    IRV & {\fontfamily{pcr}\selectfont deepchem.models.TensorflowMultitaskIRVClassifier} \\
    \hline
    GC & {\fontfamily{pcr}\selectfont deepchem.models.GraphConvModel}\\
    \hline
    Weave & {\fontfamily{pcr}\selectfont deepchem.models.WeaveModel}\\
    \hline
    DAG & {\fontfamily{pcr}\selectfont deepchem.models.DAGModel}\\
    \hline
    DTNN & {\fontfamily{pcr}\selectfont deepchem.models.DTNNModel}\\
    \hline
    ANI-1 & {\fontfamily{pcr}\selectfont deepchem.models.ANIRegression}\\
    \hline
    MPNN & {\fontfamily{pcr}\selectfont deepchem.models.MPNNModel}\\
    \hline
    \end{tabular}
    \begin{tablenotes}
        \item {$^a$} These models are based on scikit-learn package.\cite{sklearn}
        \item {$^b$} XGBoost is based on xgboost package.\cite{xgb}
    \end{tablenotes}
    \label{tab:deepchem_command}
\end{table}

\subsection{Model validation}
MoleculeNet includes multiple models that are previously proposed. To validate our reimplementation, here we compare the performances of our implementation with reported values in previous papers. All model validation scripts and trained models can be found in DeepChem.

Note that performances of our models might be different from values in the benchmark tables due to no limitation imposed on running time(more epochs), different random splitting patterns, etc.

\subsubsection{Graph Convolutional models}

We evaluate the model on ESOL dataset, note that we provide performances based on a 80/10/10 random train, valid, test splitting, while the original paper reported performance under cross validation.\cite{graphconv_feat}
~\\\\
RMSE in logS(log solubility in mol per litre):
\begin{itemize}
    \item Original result: $0.52\pm0.07$
    \item Reimplementation: $0.39$ for valid subset, $0.31$ for test subset
\end{itemize}

\subsubsection{Directed Acyclic Graph models}

We evaluate the model on ESOL dataset with the same splitting pattern, the original paper reported performance under 10-fold cross validation.\cite{lusci2013deep}
~\\\\
RMSE in logS(log solubility in mol per litre):
\begin{itemize}
    \item Original result: $0.58\pm0.07$
    \item Reimplementation: $0.68$ for valid subset, $0.58$ for test subset
\end{itemize}

\subsubsection{Weave models}

We evaluate the model on Tox21 dataset, using 80/10/10 random train, valid, test splitting. The original paper reported performance as median score of 5-fold cross validation.\cite{kearnes2016graphconv}
~\\\\
mean ROC-AUC:
\begin{itemize}
    \item Original result: $0.846\sim0.867$ for different model structure settings.
    \item Reimplementation: $0.857$ for valid subset, $0.843$ for test subset
\end{itemize}

\subsubsection{Deep Tensor Neural Network}

We evaluate the model on the atomization energy task of qm9, using 80/10/10 random train, valid, test splitting.(train subset with 106,400 samples) The original paper reported performance using different size of training set.\cite{schutt2016quantum}
~\\\\
MAE in kcal/mol:
\begin{itemize}
    \item Original result: $0.93\pm0.02$ with 2 DTNN layers and 100,000 training samples.
    \item Reimplementation: $1.15$ for valid subset, $1.26$ for test subset
\end{itemize}

\subsubsection{Message Passing Neural Network}

We evaluate the model on the HOMO-LUMO gap task of qm9, using 80/10/10 random train, valid, test splitting.(train subset with 106,400 samples) The original paper reported performance with a training set containing 110,462 randomly picked samples.\cite{MPNN} Due to that no hyperparameter is specified for the model, we are not able to fully repeat the results.

Note that the original paper trained a single model for each task in the qm9 dataset. Here we only picked one representative task to compare. 
~\\\\
MAE in eV:
\begin{itemize}
    \item Original result: $0.0544$
    \item Reimplementation: $0.0997$ for valid subset, $0.101$ for test subset
\end{itemize}

\subsubsection{Influence Relevance Voting}

We evaluate the model on the HIV dataset, using 80/10/10 random train, valid, test splitting. The original paper reported performance under 10-fold cross validation.\cite{IRV}
~\\\\
ROC-AUC:
\begin{itemize}
    \item Original result: $0.845$
    \item Reimplementation: $0.840$ for valid subset, $0.852$ for test subset
\end{itemize}

\bibliography{sample}

\end{document}